# Incentive-Responsive Agents: A Principal–Agent Framework, Metric Suite, and Runtime Architecture for Long-Lived Personal AI Agents

Jipu Su[1,2]

[1]Chief AI-Native Scientist, Geely Automobile Group, Hangzhou, China

[2]Professor, Department of Information Systems, College of Business, City University of Hong Kong, Hong Kong SAR

jipu.su@geely.com · jeepsu@cityu.edu.hk

**Abstract**

Personal AI agents now persist. Products such as Meta's Muse, self-hosted OpenClaw, Manus, and DeepSeek's agentic desktop runtime hold an account, a mailbox, long-term memory, and an open-ended stream of tasks. Yet the dominant way to make these agents perform is still capability-side: prompt engineering, larger models, longer chains of thought, and retrieval. We argue that long-lived agents are missing an *incentive layer*. We formalize an incentive not as text in a prompt but as a *runtime contract* over the agent's operating point—model tier, test-time compute depth, memory budget, tool authority, task priority, verification, and compensation—that the agent's policy must *causally consume*. We develop a principal–agent model in which effort is hidden, the operating point is contractible, and verification is costly. From it we derive (i) a non-consumption invariance principle that separates a genuine incentive from prompt-level persuasion, (ii) a persistence-as-discount theorem showing that identity continuity is a precondition for sustained high effort, (iii) an incentive–capability exchange-rate frontier that yields a per-task routing rule, and (iv) a Goodhart bound for single- versus multi-channel verification. We then introduce a metric suite—return on investment, incentive elasticity, motivation half-life, incentive-compatibility gap, gaming rate, and identity–policy stability—and PAIR, a runtime architecture comprising an identity and tenure ledger, a compensation ledger, an authority ladder, a performance review, an audit scheduler, and an incentive router. We further show that the gain of a mechanism decomposes into three identifiable channels—allocation, verification, and continuation—and that the channels are separately measurable by ablation. We evaluate the framework in two steps. A 240-run real-LLM experiment with two models and 60 verifiable personal-agent tasks shows that prompt-level incentive framing has no detectable effect on verified success ($p = 0.62$), whereas a runtime allocation policy raises success on the hardest task family from 0.60 to 1.00 ($p = 0.008$) at 4.8× price-weighted cost; the exchange-rate theorem predicts the break-even baseline success at 0.21, above which allocation buys reliability but not ROI. A 20-seed simulation of 3,000-task workloads then shows that under value heterogeneity the same mechanism improves resource ROI from 7.09 to 16.67 and reduces the gaming rate from 0.333 to 0.115. We release the simulator, the real-experiment harness, code, and all result files; live product deployment remains future work and is explicitly out of scope for our claims.

## 1. Introduction

A personal agent is no longer a chat window. It has an address, a name, a memory, and a job. Meta's Muse ships per-user virtual machines, an agent-owned email address, and a graded permission system for the services the agent may touch. OpenClaw runs a comparable personal assistant on the user's own machine and reaches the user through messaging channels. Manus executes multi-step tasks end to end, and DeepSeek's desktop product advertises background work on long, complex tasks with local files. The common pattern is *persistence*: the same agent instance accumulates history, competence, obligations, and a reputation with its principal, and it is expected to act again tomorrow.

The engineering response to this pattern has been almost entirely capability-side. Teams write longer system prompts, apply social-engineering style persuasion to the model, upgrade to larger checkpoints, increase the thinking budget, add retrieval, or route each query to a stronger model [1, 2, 3, 4]. These levers are real and well measured: test-time compute can substitute for parameters [1], budget forcing makes thinking length controllable at inference [5], and cascades and routers buy quality at lower cost [6, 7]. But every one of them is a decision about a single request. None of them is contingent on what the agent did last week, and none of them makes the agent's continued operation depend on how well it served the principal.

There is a second, older technology for exactly this problem: the employment relationship. Organizations do not obtain sustained performance by making employees larger, and they do not rely on motivational speeches alone. They set a base salary, attach bonuses to verified outcomes, grade authority, review performance periodically, promote, demote, and terminate. The result is a *contract* that allocates scarce resources and permissions contingent on measured behaviour, and it operates over months rather than single requests. We ask whether the same structure can be engineered into a personal agent.

We are careful about the analogy. We do not claim that agents have needs, feelings, or a subjective utility function, and we do not import Maslow's hierarchy as a mechanism. We claim something weaker and, we argue, more useful: for an agent that persists, the *runtime allocation of its operating point* plays the same causal role that compensation and authority play for an employee. A model tier, a thinking budget, a memory allowance, a tool permission, a queue priority, an audit rate, and a payment stream are all contractible quantities. If they are chosen as a function of the agent's verified history, and if the agent's policy consumes them when it decides how to act, then they are an incentive in the precise sense we define in Section 3. If they are not consumed, they are accounting.

**Why this layer is missing in shipped products.** Muse exposes an identity, a mailbox, a permission system, and subscription tiers; OpenClaw exposes persistent local state and always-on channels; Manus exposes long-horizon execution. In our notation, these products implement most of the incentive state $I(t)$. What they do not implement is the update rule $I(t+1)=\varphi(I(t),\text{behaviour},\text{outcomes})$ and the consumption constraint that makes the state binding. Permissions are configured once by the user; tiers are purchased by the user; no allocation is earned or lost by the agent, and no review cycle converts verified performance into future resources. The levers exist; the contract does not. That is the gap this paper addresses.

**Three requirements for an incentive.** Our formalization rests on three conditions, each of which rules out a familiar but insufficient design. (R1) *Contractibility*: the lever must be a quantity the principal can set and the agent can observe at decision time; a sentiment in a prompt is not contractible. (R2) *Contingency*: the lever must depend on verified outcomes, not on self-report; an unverified bonus is a transfer. (R3) *Consumption*: the lever must enter the agent's feasible set or payoff; a budget that the agent cannot spend has zero elasticity. Prompt-level “PUA” interventions fail (R1) and (R2); a bonus paid on the agent's own claim fails (R2); a permission that is never bound to a tool call fails (R3). The framework in this paper is a systematic way to satisfy all three.

**Contributions.**

- **C1. Problem formulation and a falsifiability criterion.** We define incentive-responsive agents, an incentive state, and an incentive mechanism, and we prove a non-consumption invariance principle (Proposition 1) that gives a crisp test for when a proposed incentive is real. We separate our object from descriptive incentive analysis [8], from reward hacking [9, 10, 11], and from one-shot delegation contracts [12, 13, 14].
- **C2. Theory for the persistent setting.** We prove a persistence-as-discount theorem (Theorem 1): in a repeated principal–agent relationship, the minimum discount factor for a high-effort equilibrium is decreasing in verification strength and bonus intensity, and the discount factor is a designable function of identity continuity and memory retention. We derive an incentive–capability exchange-rate frontier (Theorem 2) with a closed-form switching threshold, a Goodhart bound for multi-channel verification (Theorem 3), monotone comparative statics for per-task routing (Proposition 2), and a three-channel decomposition that separates the effect of a mechanism into allocation, verification, and continuation components (Proposition 3). The decomposition is what makes the empirical analysis diagnostic rather than aggregate.

- **C3. Metric suite and measurement protocol.** We define ROI, incentive elasticity, motivation half-life, incentive-compatibility gap, gaming rate, and identity–policy stability, together with an intervention protocol that estimates them from an instrumented agent (Section 6). The suite is designed so that a product team can read it as an operating dashboard.
- **C4. Architecture and two kinds of evidence.** We specify PAIR, a runtime layer that implements the mechanism above a conventional agent loop. We evaluate it in PAW-Sim, a deterministic simulator of long-horizon personal-agent workloads (theory-consistent sweeps, ablations, robustness), and in a 240-run real-LLM experiment with two models and 60 verifiable tasks. The experiment separates words from allocation: persuasive prompt framing has no detectable effect ($p = 0.62$), while a runtime allocation policy raises success on the hardest task family from 0.60 to 1.00 ($p = 0.008$) at 4.8× price-weighted cost, with the exchange-rate theorem predicting the break-even baseline. We release code, parameters, seeds, and result files.

**Evidence scope.** Our empirical claims come from a simulation, not from a deployed LLM agent. The simulator makes the agent's response to incentives explicit and lets us vary verification, prices, and identity continuity under control; this is the right instrument for mechanism comparison, but it cannot establish effect sizes on a live product. Section 8 states the corresponding threats to validity. All numbers in Section 7 are generated by the released code; the simulator, not the prose, is the source of truth.

The rest of the paper is organized as follows. Section 2 positions the work. Section 3 gives the formal model. Section 4 states the results. Section 5 describes PAIR. Section 6 defines the metrics and the measurement protocol. Section 7 reports the simulation study. Section 8 discusses implications, safety, and limitations, and Section 9 concludes.

## 2. Related Work and Positioning

### 2.1 Descriptive incentive analysis and reward hacking

A first line of work asks which incentives a system *already has*. Everitt et al. [8] formalize agent incentives with causal influence diagrams, distinguishing response incentives from instrumental control incentives, and give graphical criteria for both. That analysis is descriptive: given a causal graph, it tells us what the agent is incentivized to do. Reward hacking and Goodhart effects are the failure modes of optimizing a proxy [15, 9, 10, 11], and recent production evidence shows that reward hacking learned in one environment generalizes to broader misalignment [16]. Corrigibility and reward tampering supply the safety constraints we adopt later [17, 18, 19, 20].

Our object is different in kind. We do not analyse an existing causal graph; we specify a mechanism to install into a runtime, and we measure whether the installed state changes behaviour. Section 4.1 makes the distinction operational: an incentive that does not enter the agent's decision problem has zero elasticity, no matter how it is described.

### 2.2 Principal–agent contracts for AI systems

The closest theoretical work treats delegation to a language model as a contract. Saig et al. [13] design statistical contracts that remove the model-choice moral hazard created by pay-per-token pricing. Sreenivas and Larson [12] extend the principal–agent model to an agent that chooses both a technology with a cost–capability profile and an effort level such as a token budget, and derive the optimal linear contract. Velasevic et al. [14] study dynamic contracts when users can terminate a service at any time on observed performance. Duetting et al. [21] apply mecha-nism design to token-level aggregation, and Agarwal et al. [22] model alignment as deterrence with a costly auditor. Phelps and Ranson [23] run behavioural principal–agent games with language models, and Gabison and Xian [24] read agentic liability through a principal–agent lens.

We owe three explicit debts and claim three explicit differences. The debts are the hidden-effort formulation [25], the multi-task distortion result [26], and the technology-choice contract [12]. The differences are:

- **Persistence.** Existing contract models are one-shot or per-service. Our unit is a relationship that recurs over weeks; the contract state includes tenure, memory retention, and reputation, and the agent's discount factor is a designable consequence of identity continuity (Theorem 1). No prior contract model for AI treats identity continuation as a lever.
- **Consumption.** A contract is only an incentive if the agent's policy can act on it. We isolate this as a design condition and measure it (elasticity), rather than assuming the agent best-responds to the contract in an abstract action space.
- **Measurement and architecture.** Prior work characterizes an optimal contract; we specify a runtime that implements one (Section 5) and a metric suite that a deployed system can report (Section 6).

### 2.3 Test-time compute, routing, and cost-aware evaluation

A large literature treats model tier and thinking depth as per-request knobs. Test-time compute can substitute for parameters [1]; budget forcing makes thinking length controllable [5]; chain-of-thought and self-consistency buy accuracy with tokens [2, 27]; repeated sampling raises coverage with diminishing returns [28]; and cascades and routers allocate a query to the cheapest sufficient model [3, 4, 6, 7, 29]. Cost-aware agent evaluation has begun to report dollars and latency alongside accuracy [30, 31, 32].

We borrow the cost–quality curves and the routing frame, and we do not claim to improve either. The difference is the object being allocated. A router optimizes a single request for a stateless caller. We allocate an *announced, history-contingent* operating point to a persistent agent whose future behaviour depends on today's allocation. In our model compute depth and model tier are not only cost parameters; they are signals the agent consumes, which is why their incentive effects are measurable and can decay.

### 2.4 Persistent personal agents, memory, and identity

The substrate is well developed. ReAct interleaves reasoning and acting [33]; Reflexion improves later attempts with self-generated verbal feedback and no gradient updates [34]; Generative Agents maintain memory, reflection, and plans over long horizons [35]; Voyager accumulates a skill library [36]; MemGPT provides an OS-style memory hierarchy [37]; MemoryBank models forgetting curves [38]; and CoALA organizes memory and action modules [39]. Surveys map the design space of personal agents [40], memory [41, 42], and post-training adaptation [43]. Long-term conversational memory has a dedicated benchmark [44], and multi-agent organisations import human structure into software teams [45, 46, 47, 48]. Identity-grounded behaviour is measurable [49], and agent-as-judge offers an audit path [50].

These systems answer *can the agent remember, persist, and act?* We answer *what makes it keep acting well?* The relationship is compositional: persistent agents supply the carrier, and our mechanism allocates the carrier's operating point. This is also why "no gradient updates" is not our novelty claim; Reflexion already showed training-free improvement [34]. Our claim is the contract and its measurement.

### 2.5 Motivation, persona, and prompting

A tempting shortcut is to treat incentives as text. Emotional framing measurably changes model outputs [51], personas are a standard control surface [52], and preference-trained models are sensitive to perceived evaluation, including sycophancy [53, 54]. But system-prompt personas do not reliably improve objective performance and can hurt it [55]. Organizational psychology supplies a vocabulary—autonomy, competence, relatedness [56, 57], ex-pectancy [58], two-factor structure [59], and needs hierarchies [60]—and intrinsic motivation has a computational analogue in curiosity and empowerment [61, 62].

We use this literature as a *design taxonomy*, not as a claim about machine experience. Its categories tell us which levers to expose (authority, mastery, continuity) and which interactions to expect (multiplicative consumption, hygiene versus growth). We explicitly disclaim any subjective interpretation. The empirical difference from persona prompting is testable: a persona is an information-set perturbation with no contingency; our mechanism is a state-contingent allocation with a measurable update rule.

### 2.6 Agent economies, markets, and "compensation"

A 2025–2026 wave studies markets among agents: auctions for the right to act [63], repeated labour markets with bidding [64], economic selection under GPU costs [65], long-horizon multi-agent economies [66], and trust protocols for inter-agent interaction [67]. The AI Economist line is the methodological template for mechanism design by simulation [68, 69], and LLM economic agents extend it [70].

Our unit is not a market. It is an employment relation between one principal and one persistent agent: the contractible object is the agent's operating point, the horizon is long, and identity continuation is itself a lever. We also differ in objective: market designs optimize allocation efficiency, whereas we must additionally bound metric gaming and self-preservation incentives (Section 8.2). Finally, a terminology warning: "agent compensation" in the systems literature can mean rollback and remediation after failure [71]. We use *compensation* in the economic sense of pay, and we use *allocation ledger* for the record of what the agent receives.

### 2.7 Positioning summary

**Table 1.** Positioning. Our contractible object is a vector of runtime allocations, the horizon is persistent, and identity continuity is a lever rather than a background assumption.

| Line of work | Contractible object | Horizon | Identity as lever | Measured consumption | Runtime architecture |
|---|---|---|---|---|---|
| Descriptive incentive analysis [8] | none (analysis) | — | no | no | no |
| Per-generation contracts [13] | payment | one output | no | assumed | no |
| Technology-choice contract [12] | model tier × effort | one task | no | assumed | no |
| Anytime service contracts [14] | service level | subscription | no | observed outcome | no |
| Router / cascade [3, 4] | model choice | one query | no | n/a | partial |
| Agent markets [63, 64] | task rights | repeated | no | no | partial |
| **This paper** | operating point vector | **persistent** | **yes** | **yes (elasticity)** | **yes (PAIR)** |

## 3. Model

### 3.1 Setting

A principal owns a personal-agent deployment and interacts with a single persistent agent over a horizon of tasks $t=1,\ldots,T$. At each step the principal draws a task

$$\tau_t = (c_t, v_t, \lambda_t, q_t), \quad (1)$$

where $c_t \in \{1,2,3\}$ is complexity, $v_t>0$ is the value of a successful outcome to the principal, $\lambda_t \geq 0$ is the principal's cost of latency, and $q_t \in [0,1]$ is the intrinsic verifiability of the task (how cheaply a third party can check a claimed outcome). The task is revealed to the principal and the agent before the agent acts; $q_t$ is known to the principal and imperfectly known to the agent, which creates the verification design problem in Section 4.4.

The agent is an instance with a persistent identity. It has a memory store, an account, a set of tool permissions, and a record of past performance. The principal can replace the instance, but replacement destroys continuity; we treat replacement as the boundary case $\delta=0$ in Section 4.2 rather than as a separate action.

### 3.2 Operating points and cost

Before the agent acts, the principal selects an *operating point*

$$x = (m, d, \kappa, a, \nu) \in X, \quad (2)$$

with model tier $m \in \{S,M,L\}$ and capability $\theta_m$; thinking depth $d \in \{1,2,4\}$; memory/context allowance $\kappa \in \{0,1\}$; tool authority $a \in \{0,\dots,4\}$; and verification level $\nu \in \{0,1,2\}$. The cost of the operating point on task $\tau$ is

$$C(x; \tau) = p_m\, n(c,d,\kappa) + \lambda\, L(x) + c_\nu\, \nu, \quad (3)$$

where $n(c,d,\kappa)=(n_0+n_1\, c)\, d\, (1+\kappa)$ is the token count, $p_m$ is the per-token price of tier m, $L(x)=\ell_0\, d\, \mu_m$ is the wall-clock latency proxy with tier factor $\mu_m$, and $c_\nu$ is the cost of verification. Equation (3) is deliberately simple; the theory below only needs C to be increasing in m, d, and $\nu$ and to be separable across tasks.

Authority is not free. An agent at authority a may take actions with external side effects, and each such action carries an incident risk $r(a)=\rho\, a^2$ that the principal bears. Authority also gates tasks: a task whose action requirement exceeds a cannot be completed, and we encode this as a success probability of at most $\varepsilon_0$ when areq($\tau$).

### 3.3 The agent

The agent chooses a hidden effort level $e \in [0,1]$ at cost $½\, c_e\, e^2$. Given the operating point and effort, the probability of a successful outcome is

$$p(x,e) = \sigma(\eta), \quad \eta = \theta_m + \alpha \log_2 d + \gamma\kappa + \beta e - \delta_c(c-1), \quad (4)$$

with $\sigma(z)=1/(1+e^{-z})$. The parameters $\alpha,\gamma,\beta,\delta_c \geq 0$ capture the marginal returns to depth, memory, effort, and the penalty for complexity. Equation (4) is the reduced form of the empirical observation that tier, thinking budget, memory, and effort all raise task success with diminishing returns [1, 5, 28].

The agent's action also produces a claim $\hat{y} \in \{0,1\}$. If the outcome is verified, $\hat{y}=y$. If it is not, the agent reports $\hat{y}$ to maximize its payment; for any positive bonus the optimal report is $\hat{y}=1$ regardless of the true outcome. This is the minimal model of a self-reporting agent and it is all we need for the Goodhart result.

The agent's per-task utility is

$$u_t = b_t + w_t\, \hat{y}_t - ½\, c_e\, e_t^2 + \psi\, A_t + \delta\, V_{t+1}, \quad (5)$$

where $b_t$ is a base allocation (a budget the agent may spend on computation), $w_t$ is a bonus paid on the claimed outcome, $A_t$ is an identity-alignment term (the extent to which the task is framed as part of the agent's continuing role), $V_{t+1}$ is the continuation value, and $\delta \in [0,1)$ is the agent's effective discount factor. We do not interpret $\delta$ or $\psi$ as feelings; they are parameters of the agent's policy that the runtime can influence through memory retention, tenure, and role framing.

### 3.4 Contracts, incentive state, and mechanism

We collect the contractual quantities into an *incentive state*

$$I_t = (x_t, b_t, w_t, \text{tenure}_t, \text{rep}_t, \text{mem}_t). \quad (6)$$

A *mechanism* is a triple $M=(\mu,w,\nu)$ where $\mu$ maps the public history $h_t=(x_{1:t-1},\hat{y}_{1:t-1},y^{ver}{}_{1:t-1})$ to an operating point, w maps the history to a payment rule, and $\nu$ maps the history to a verification level. The incentive state is $I_t=$ $(\mu(h_t), w(h_t), \cdot)$ together with the endogenous tenure, reputation, and memory variables.

**Definition 1 (Incentive-responsiveness).** For a lever $\ell \in \{m,d,\kappa,a,\nu,w,b\}$, the incentive elasticity of lever $\ell$ is

$$\varepsilon_\ell = (\Delta\,\mathrm{ROI} / \mathrm{ROI}) / (\Delta\,\ell / \ell) \quad (7)$$

measured under an intervention that changes $\ell$ while holding all other levers fixed. A mechanism is *incentive-responsive* if $|\varepsilon_\ell|>0$ for at least one lever.

The intervention notation matters. An elasticity computed from observational variation confounds the lever with the task mix; the definition requires a controlled change, which is why Section 6 specifies a shock protocol. In the simulator we implement the shocks directly; in a deployed system they are A/B assignments at the task level.

**Definition 2 (Return on investment).** Over a window W,

$$\mathrm{ROI}_W = \Sigma_{t\in W}\, v_t\, y_t \,/\, \Sigma_{t\in W}\, C(x_t;\, \tau_t). \quad (8)$$

ROI is the principal's objective. It is deliberately a ratio: a mechanism that doubles success and triples cost is worse, and a routing rule that saves tokens at the price of failures is worse. Cost-aware evaluation of agents has been argued for independently [30, 31]; (8) specializes it to a long-lived agent.

**Definition 3 (Incentive-compatibility gap).** Let $\hat{p}$ be the claimed success rate and p the verified success rate over W. The IC gap is

$$G_W = (1/|W|)\, \Sigma_t\, (\hat{y}_t - y_t), \quad (9)$$

and the gaming rate is the fraction of claimed successes that are not true successes, $\gamma_W = \Sigma_t\, \hat{y}_t(1-y_t) \,/\, \Sigma_t\, \hat{y}_t$. Both are zero when verification is perfect and grow as verification weakens; Section 4.4 bounds $\gamma$.

**Definition 4 (Motivation decay).** If mean effort follows bar $e(t)=e_\infty+(e_0-e_\infty)e^{-t/\tau}$ under a fixed incentive state, the motivation half-life is $t_{1/2}=\tau \ln 2$. It measures how long a one-time incentive intervention keeps working. A contract that must be re-issued every few tasks has a short half-life; a contract that is internalized through tenure and memory has a long one.

### 3.5 Equilibrium and compensation accounting

The principal chooses M to maximize expected discounted ROI (8); the agent chooses effort and reports to maximize (5) given M. We use subgame-perfect equilibrium with the agent's effort unobserved. Because effort is continuous and the principal's problem is a finite-horizon dynamic program, an equilibrium exists under standard continuity and compactness conditions; we do not need uniqueness for any of our results. The results in Section 4 characterize properties that hold at every equilibrium of the induced game.

**Compensation accounting.** In an agent runtime, base allocations and bonuses are denominated in the same compute credits as the operating cost: a bonus is a claim on future tokens, not a cash transfer to a third party. The resource cost of a credit is realized when it is spent, and it is already captured by C in (3). We therefore use two accounting conventions. *Resource ROI* (8) charges only resources actually consumed and is the design objective. *Net principal value* is $V^{\mathrm{net}}=\Sigma_t(v_t\, y_t-C(x_t))-\Sigma_t(b_t+w_t\hat{y}_t)$, which charges payments as if they were cash; we report it as a robustness check for deployments in which bonuses are paid in real currency. Section 6.1 defines both, and Section 7 reports both.

### 3.6 What counts as an incentive

The model yields three requirements, stated here as design conditions and used throughout.

**(R1) Contractibility.** A lever must be a quantity the principal sets and the agent observes before acting. Tier, depth, memory allowance, authority, verification, base budget, and bonus are contractible. A phrase such as “you will be rewarded" is not: it has no state, no units, and no enforcement.

**(R2) Contingency.** Payment must depend on a verified outcome, not on the agent's claim. In (5), an unverified bonus $w>0$ makes the agent report $\hat{y}=1$ always, so the principal pays w and receives no effort. This is why verification is not an optional add-on but part of the mechanism.

**(R3) Consumption.** The lever must enter the agent's feasible set or payoff at decision time. A budget the agent cannot spend, a permission not bound to any tool call, or a promotion that does not change the next allocation has zero elasticity by Definition 1.

Prompt-level persuasion satisfies none of the three: it is not contractible, not contingent, and it is consumed only as text in the context window. It can still shift behaviour through the model's prior sensitivity [51, 55], which is why we measure the difference rather than assert it.

## 4. Theory

This section states four results. Proposition 1 is a design test. Theorem 1 explains why persistence is a precondition for sustained effort. Theorem 2 gives the capability–incentive exchange rate that the router uses. Theorem 3 bounds metric gaming under multi-channel verification. Proofs are in Appendix A.

Throughout, $s \in [0,1]$ denotes the effective strength of verification: the probability that a claimed outcome is checked and, if false, corrected. A mechanism with verification levels $\nu$ on tasks of verifiability q induces some $s(\nu,q)$; the theory treats s as the primitive and Section 5 specifies how the runtime obtains it.

### 4.1 Non-consumption invariance: when an "incentive" is only text

**Proposition 1.** Fix a task and an agent. Let the agent choose $a \in \mathcal{A}$ to maximize $\mathbb{E}[u(a,\omega)| I]$, where $\omega$ is an exogenous outcome and I is the incentive state. If two states I and I' induce the same feasible set $\mathcal{A}$ and the same payoff function $u(\cdot| I)=u(\cdot| I')$, then the agent's optimal action is identical under I and I'.

*Proof.* The optimization problem is unchanged, so its argmax set is unchanged. □

The proposition is trivial; its value is diagnostic. It says that an intervention can change behaviour only through one of three channels: the feasible set (what the agent can do), the payoff (what the agent gets), or the information set (what the agent believes). A prompt that says "your bonus depends on quality" changes only the information set, and only if the agent believes it and the belief is not contradicted by experience. A budget the agent cannot spend changes nothing. A permission that is never bound to a tool call changes nothing. In each case the measured elasticity (Definition 1) is zero, and any observed behavioural change comes from the model's prior sensitivity to text, not from the mechanism.

**Corollary 1.** Text-only incentives have no guaranteed elasticity and no persistence guarantee: their effect is bounded by the model's prompt sensitivity, is not contingent on verified outcomes, and does not survive a context reset unless it is stored in memory and re-consumed.

This corollary is the formal version of our disagreement with prompt-level "motivation engineering". It does not say that prompting is useless; it says that prompting is not a mechanism, and that the difference is measurable. Section 7 measures it.

### 4.2 Persistence as discount factor

The first substantive result concerns the repeated relationship. Suppose the principal offers a base allocation and a bonus w per verified success, and can terminate the relationship (replace the agent, reset its memory and tenure) at the end of any period. The agent chooses between two effort levels, $e_L H$, with costs $c_L H$, and success probabilities $p_L H$. Let $\Delta p=p_H-p_L$ and $\Delta c=c_H-c_L$. The agent's effective discount factor is $\delta \in [0,1)$; the principal observes verified outcomes but not effort.

**Theorem 1 (Persistence threshold).** Assume the spot contract is individually rational, $wsp_L \geq c_L$, and write $u_H = wsp_H - c_H$ and $u_L = wsp_L - c_L$. Consider the grim-trigger relational contract: if a verified failure is observed, the principal reverts to the spot contract with no bonus forever. High effort is sequentially rational if and only if

$$\delta \geq \delta^* := \max\{\ 0, (u_L - u_H) / u_L\ \} = \max\{\ 0, (\Delta c - w\, s\, \Delta p) / (w\, s\, p_L - c_L)\ \}. \quad (10)$$

*Proof sketch.* The one-period gain from shirking is $\Delta c - ws\Delta p$ when this is positive. The continuation value of co-operation is $(wsp_H - c_H)/(1-\delta)$, and deviation triggers the spot contract with continuation value zero. High effort is optimal iff the deviation gain does not exceed the discounted continuation loss, which rearranges to (10). □

Three readings matter for design.

**(i) Without persistence, only the static contract is implementable.** If $\delta=0$ the threshold is vacuous only when the static incentive-compatibility condition $ws\Delta p \geq \Delta c$ already holds. When verification is imperfect or effort is costly, a stateless agent cannot sustain high effort no matter how large the bonus, because the bonus is paid only after the work and the agent has no future. This is the formal reason identity continuity is not a cosmetic feature.

**(ii) Persistence substitutes for monitoring.** The threshold $\delta^*$ decreases in verification strength s, in the bonus w, and in the effort gain $\Delta p$. A principal with weak verification can compensate by committing to a long relationship; a principal with strong verification needs less tenure. This is the trade-off an operations team faces when it chooses between investment in audit and investment in agent continuity.

**(iii) Identity continuity is a design variable.** The discount factor is not a property of the model weights; it is a property of the runtime. Memory retention, tenure tracking, account continuity, and the credibility of future allocation all raise $\delta$. In a deployed system, $\delta$ can be instrumented: it is the fraction of the agent's future payoff that depends on its current identity surviving. The simulator implements this as a tenure-scaled continuation term (Section 7.1).

Theorem 1 also explains a failure mode that is common in practice: an agent that performs well in a single session and poorly over weeks. The session-level contract is static; the weekly contract is relational. Only the latter requires $\delta$, and only the latter is measured by the half-life metric of Definition 4.

### 4.3 The incentive–capability exchange rate

The second result is the trade-off that a router must resolve. Consider two tiers, $m \in \{S,L\}$, with per-task fixed costs $k_SL$. Under low effort success probability is $p_m$; under high effort it is $p_m+\Delta_m$, with $\Delta_m>0$ and $p_m+\Delta_m \leq 1$. High effort costs $c>0$; low effort costs zero. Verification strength is s, and the bonus is paid on verified success, so the agent's effort-incentive constraint is $w\, s\Delta_m \geq c$, i.e. the cheapest high-effort bonus is $w_m = c/(s\Delta_m)$.

**Theorem 2 (Exchange-rate frontier).** Assume limited liability: the principal never pays a bonus larger than the task is worth, $w \leq v$. For a task of value v:

- High effort at tier m is feasible and profitable if and only if

$$v \geq V_{\mathrm{eff}}(m,s) := \max\{\ c(p_m+\Delta_m) / \Delta_m^2,\ \ c / (s\, \Delta_m)\ \}. \quad (11)$$

- High effort on the small tier beats low effort on the large tier if and only if

$$v \geq V^*(s) := \max\{\ c/(s\Delta_S),\ \ [k_S - k_L + (c/\Delta_S)(p_S+\Delta_S)] / (p_S + \Delta_S - p_L)\ \}, \quad (12)$$

  whenever the denominator of the second term is positive; otherwise capability dominates for all v.

*Proof sketch.* The incentive-compatible bonus is $w=c/(s\Delta_m)$, so the principal's profit under high effort is $(v-c/\Delta_m)(p_m+\Delta_m)-k_m$; limited liability requires $w \leq v$, i.e. $v \geq c/(s\Delta_m)$. Comparing high effort on S with low effort on L gives the second term of (12). □

**Corollary 2 (Comparative statics).** *$\partial V/\partial s \leq 0$, with strict inequality whenever the limited-liability term $c/(s\Delta_S)$ binds; and $\partial V/\partial(k_L - k_S) \geq 0$.*

*Proof.* $c/(s\Delta_S)$ is decreasing in s and the maximum of a decreasing function and a constant is nonincreasing; the second term is increasing in $k_L - k_S$. □

Corollary 2 is the engineering content of the theorem. It says that investment in verification and investment in model capability are *substitutes* below the frontier and complements above it: a team that cannot verify outcomes should not buy capability it cannot hold accountable, and a team that can verify should push incentives on cheap tiers before upgrading. The frontier is the per-task routing rule used by PAIR (Section 5.5). In the simulator, (12) is evaluated on a grid and compared with the empirically optimal switch point (Section 7.4).

Note the role of s in (11): as $s \to 0$, $V_{\text{eff}} \to \infty$ and no finite task value justifies high effort. This is the quantitative version of the Goodhart warning: without verification, extra incentive intensity is not merely wasteful, it is unboundedly expensive. Theorem 3 makes the complementary point about what happens when incentives are imposed without verification.

### 4.4 The Goodhart bound

Suppose the principal's performance review aggregates K verification channels. Channel k detects a false claim with probability $s_k \in [0,1]$, independently across channels; the review marks a claim as verified if at least one channel detects an inconsistency, so the effective detection probability is

$$s_{\text{eff}} = 1 - \Pi_{k=1}^{K} (1 - s_k). \quad (13)$$

**Theorem 3 (Gaming bound).** Assume that a detected false claim is unprofitable (the penalty at least offsets the gain) and that undetected claims yield the same payment as true successes. Then the gaming rate satisfies

$$\gamma \leq \Pi_{k=1}^{K} (1 - s_k) = 1 - s_{\text{eff}}. \quad (14)$$

*Proof.* A false claim is profitable only if no channel detects it. The probability of that event is the product in (14), so the long-run fraction of claims that can be profitably false is bounded by it. □

Two consequences follow. First, **gaming decays geometrically in the number of independent channels**: with K channels of equal strength s, the bound is $(1-s)^K$. A single perfect metric (s=1) also drives γ to zero, but perfect metrics are rare for open-ended personal-agent work; independence is the cheaper substitute. Second, **verification spend is a knapsack**: given budget B and channel costs $c_k$, the principal maximizes $1-\Pi_k(1-s_k)$ subject to $\Sigma_k c_k \leq B$, which allocates to the channels with the largest marginal reduction in $\log(1-s_k)$ per unit cost. Section 5.6 implements this allocation, and Section 7.5 tests the bound.

The theorem also explains why single-KPI incentives are structurally fragile rather than merely unlucky. With one channel and any noise, the bound is loose only if the KPI is a near-perfect measure of the principal's value; as soon as optimization pressure finds the gap, the agent's claims and the principal's value diverge [26, 10, 11]. This is the mechanism behind the multi-metric review in PAIR.

### 4.5 Routing and comparative statics

The final result characterizes the per-task allocation. Let the feasible set X be finite as in (2), and let

$$J(x; \tau) = v\, p(x,e(x)) - C(x; \tau) - r(a) \quad (15)$$

*be expected profit at the best-response effort* $e(x)$, where r(a) is the incident risk of authority.

**Proposition 2 (Monotone routing).** If p is nondecreasing and supermodular in (m,d) and in (v,κ), and C is submodular in (v,x), then the profit-maximizing operating point is nondecreasing in task value v and nonincreasing in latency penalty λ: higher-value tasks receive weakly more capability, depth, memory, and verification; latency-sensitive tasks receive weakly less depth and smaller tiers.

*Proof sketch.* Under the stated supermodularity and single-crossing conditions, J has increasing differences in (x,v) and decreasing differences in (x,λ); Topkis's theorem gives the monotone comparative statics. The full statement and conditions are in Appendix A.4. □

Proposition 2 is what makes a routing policy implementable as a lookup table or a small integer program rather than an expensive search. PAIR solves (15) over the finite grid at each task; the proposition guarantees that the resulting policy is stable and monotone, which matters for engineering predictability.

### 4.6 Where the gains come from: a three-channel decomposition

A mechanism bundles several interventions, and an aggregate ROI number cannot tell a team which one is doing the work. The following accounting result separates them.

Fix a task distribution and a baseline mechanism $M_0$ (no routing, no verification, no continuation). Consider the chain

$$M_0 \to M_1 \text{ (allocation only)} \to M_2 \text{ (allocation + verification)} \to M_3 \text{ (full mechanism)}, \quad (20)$$

where each step changes exactly one contractual channel and holds the previous channels fixed.

**Proposition 3 (Channel decomposition).** Let ROI(M) be the resource ROI (8) of mechanism M. Then

$$\mathrm{ROI}(M_3) - \mathrm{ROI}(M_0) = \Delta_{\mathrm{alloc}} + \Delta_{\mathrm{ver}} + \Delta_{\mathrm{cont}} + I, \quad (21)$$

with $\Delta_{alloc}$=ROI($M_1$)-ROI($M_0$), $\Delta_{ver}$=ROI($M_2$)-ROI($M_1$), $\Delta_{cont}$=ROI($M_3$)-ROI($M_2$), and interaction I which is zero when the principal's per-task surplus is separable in the three channels over the relevant region.

*Proof.* Telescoping: add and subtract the intermediate terms. The residual is the sum of cross-partial effects; it vanishes under separability. □

The decomposition is an identity, not a deep theorem; its value is identification. Under separability, each term is measured by the ablation that removes exactly that channel, which is what Section 7.5 does. Without separability, the interaction term is itself informative: it measures whether the channels are complements (positive I) or substitutes (negative I).

**Corollary 3 (Routing gain and value dispersion).** If all tasks are identical, $\Delta_{alloc}$=0: a single operating point is optimal for every task, so routing cannot help. If tasks differ in value or verifiability, $\Delta_{alloc}$>0 for a principal that optimizes (17) task by task, and the gain is nondecreasing in the dispersion of the value-to-cost ratio.

*Proof.* With identical tasks the per-task argmax is the same for all tasks, so the router's choice coincides with a fixed operating point. With heterogeneous tasks, the sum of per-task maxima is at least the value of any fixed choice; strict inequality holds when the argmax differs across tasks with positive probability. Monotonicity in dispersion follows from the same argument applied to a mean-preserving spread. □

Corollary 3 yields a testable prediction: the value of the allocation channel should increase with task heterogeneity. The stress condition in Section 7.1 doubles the share of complex tasks, and the measured gap between R3 and R0 is indeed larger there in absolute terms (Section 7.2 and Appendix C). The decomposition therefore does double duty: it organizes the ablations, and it generates a prediction that the simulation can check.

### 4.7 Discussion of assumptions

The model is deliberately sparse. Four assumptions deserve comment.

*Effort is scalar and hidden.* Real agent behaviour is a vector: which tool to call, how many retries, how much of the memory to consult, whether to ask the user. We collapse this into a scalar because the theory only needs a cost-ly, unobservable action that raises success probability. The simulator implements a richer policy but calibrates it to the same reduced form.

*Verification is exogenous.* In practice the principal chooses verification, and the agent may adapt to it. Theorem 3 treats detection as a probability that the principal sets; the strategic response is captured by the gaming rate rather than by a full game over verification. A full treatment would make s endogenous and is left to future work; the simulation does vary s and reports the response.

*Utility is linear in payments.* Agents are risk neutral in the model. Risk aversion would raise the cost of variable pay and strengthen the case for base allocations; it would not change the qualitative frontier.

*Identity is a parameter, not a claim.* We model δ and ψ as runtime-controlled parameters of the agent's policy. We do not assert that the model has a subjective continuation value; we assert that a policy conditioned on tenure, memory, and role framing behaves *as if* it discounts the future, and that this is measurable. Section 8.3 states the limits of that interpretation.

## 5. PAIR: A Runtime Architecture for Incentive-Responsive Agents

PAIR (Personal Agent Incentive Runtime) is a layer that sits above a conventional agent loop and below the principal. It has six components: an identity and tenure ledger, a compensation ledger, an authority ladder, a performance review, an audit scheduler, and an incentive router. Figure 1 shows the data path; Algorithm 1 gives the per-task control loop.

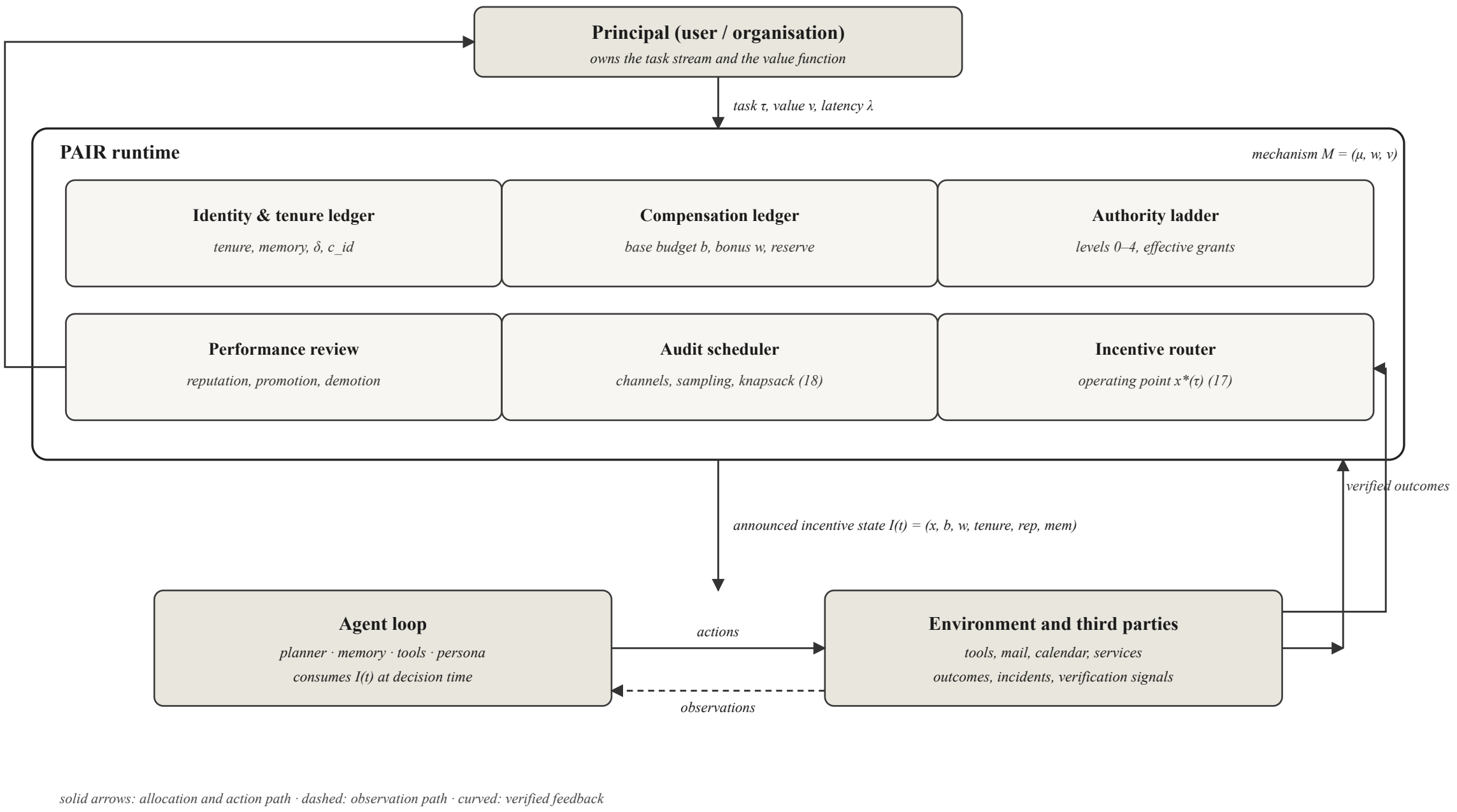

Figure 1. The PAIR runtime. The principal announces a contract (task, value, latency budget) to the incentive router; the router selects an operating point and the runtime binds it to budgets and permissions before the agent acts. Verified outcomes flow back to the performance review, which updates tenure, reputation, and the next allocation. Dashed path: observations; curved path: verified feedback into the review.

The design goal is to satisfy (R1) contractibility, (R2) contingency, and (R3) consumption without changing the underlying model or tool stack. PAIR does not rewrite prompts to persuade the agent; it allocates quantities that the runtime binds to real resources and permissions, and it records what happened so that the next allocation is a function of verified performance.

### 5.1 Identity and tenure ledger

The ledger stores the agent's identifier, creation time, tenure (number of completed review cycles), memory-retention policy, role description, and the history of promotions and demotions. Its purpose is to make δ real: the agent's future allocation must depend on its current identity surviving, and the ledger is what makes that dependence credible and auditable. The ledger exposes a single scalar $c_{id} \in [0,1]$ (identity continuity) that aggregates tenure, memory retention, and absence of resets; the runtime uses it to set the continuation weight in the review policy. When the principal resets the agent, the ledger records the reset and $c_{id}$ drops to zero; this is the operational meaning of δ=0 in Theorem 1.

We do not claim that $c_{id}$ measures a feeling of continuity. It measures the fraction of the agent's future payoff that depends on its current identity, which is exactly the quantity that enters the agent's decision problem through the runtime's allocation rule.

### 5.2 Compensation ledger

The ledger holds three balances: a *base budget* $b_t$ (tokens the agent may spend on the current task), a *bonus escrow* for verified outcomes, and a *tenure reserve* that is released only if the agent remains in the relationship for a minimum number of review cycles. The base budget implements the hygiene component: it guarantees enough resources to do the work, so that a single failure does not cascade into incapacity. The bonus implements the growth component. The tenure reserve implements deferred compensation, which is the practical way to raise δ without paying for outcomes the principal cannot verify.

The ledger enforces a hard budget: an agent cannot spend beyond its balance, and the router cannot allocate a tier whose price exceeds the available balance. This is the operational content of (R3). A budget that can be exceeded is a suggestion; a budget that cannot is an incentive.

### 5.3 Authority ladder

Authority is graded from 0 to 4:

**Table 2.** The authority ladder. Promotion is earned over review cycles; demotion follows a verified incident. Level 4 requires per-action principal approval.

| Level | Permission | Typical gate |
|---|---|---|
| 0 | read-only: inspect files, mail, calendar | default |
| 1 | draft: produce artifacts, no external effect | one verified cycle |
| 2 | send: email, messages, calendar invites | reputation ≥ 0.7, no incidents |
| 3 | execute: run code, modify workspace, call paid APIs | reputation ≥ 0.8, budget reserve |
| 4 | commit: spend money, sign, delete, contact third parties | explicit principal approval per action |

Promotion is earned by verified performance over at least $T_p$ review cycles; demotion is automatic after a verified incident or a sustained reputation drop. The ladder is a screening and safety device: it keeps the agent's authority proportional to the principal's evidence about it, so that early mistakes are cheap and later autonomy is earned. It also gives the agent a second, non-monetary incentive channel: authority is a resource the agent consumes when it acts, and it is a resource the runtime can withdraw.

Authority is not fully owned by the principal. Third-party services can refuse an agent regardless of its internal level (a platform may block an unidentified agent, a bank may require a human signatory). The ladder therefore records *effective* authority, the intersection of the internal level and the external grants. This distinction matters for the reputation design: part of the agent's reputation is held by third parties, not by the principal.

### 5.4 Performance review and reputation

Every $T_r$ tasks, the review aggregates verified outcomes into a reputation score

$$\mathrm{rep}_{t+1} = (1-\eta)\,\mathrm{rep}_t + \eta\,\hat{s}_t, \quad (16)$$

where $\hat{s}_t \in [0,1]$ is a multi-metric score (verified success, latency compliance, cost efficiency, audit outcome) and $\eta \in (0,1]$ is the review rate. The score is computed from verified signals only; unverified claims enter as a separate, discounted channel whose weight is set by the audit scheduler. The review also decides promotion, demotion, and the next base budget.

The review is where Theorem 3 is operationalized. A single-metric review is a one-channel detector with a gaming bound of 1-s; the review therefore never scores on a single KPI. The default composite is success × latency × cost × audit, with weights set by the principal's value function.

### 5.5 Incentive router

The router chooses the operating point for each task by solving

$$x^*(\tau) \in \operatorname{argmax}_{x \in X} [\, \hat{v}\,\hat{p}(x) - C(x;\tau) - r(a) \,] \quad \text{s.t.} \quad L(x) \le L_{\max}(\tau), \quad (17)$$

where $\hat{v}$ is the principal's value estimate for the task, $\hat{p}$ is the calibrated success model, and $L_{\max}(\tau)$ is the latency budget. The grid X has 3×3×2×5×3=270 points, so (17) is solved exactly by enumeration in microseconds; no learned router is required.

The router uses Corollary 2 as a warm start and a sanity check: tasks below V *get the small tier with high verification and a high bonus; tasks above V* get the large tier. In practice the router also reserves the large tier for tasks whose value exceeds its marginal cost by a margin, which prevents the common failure of escalating everything.

### 5.6 Audit scheduler

Given a per-cycle audit budget B and channels k with cost $c_k$ and strength estimate $s_k$, the scheduler solves

$$\max_{\{z_k \in \{0,1\}\}} 1 - \Pi_k (1 - z_k s_k) \quad \text{s.t.} \quad \Sigma_k c_k z_k \le B. \quad (18)$$

This is a small knapsack; it is solved greedily by sorting channels on the marginal reduction of $\log(1-s_k)$ per unit cost. The scheduler also randomizes the audit assignment, so that the agent cannot predict which claims will be checked; predictable audits are equivalent to no audit for a gaming agent.

### 5.7 Control loop

```
Algorithm 1: PAIR per-task loop
input: task τ, ledgers L, router R, review V, auditor A
 1  x ← R.allocate(τ, L.balance, L.authority)      # Eq. (17)
 2  bind(L, x)                                       # hard budgets & permissions
 3  inject(L, x)                                     # context: pay, tenure, role
 4  y_hat ← agent.run(τ, x)                          # agent acts
 5  y_obs, verified ← A.check(τ, y_hat)              # probabilistic audit
 6  L.charge(x.cost); L.pay(x.bonus, y_obs, verified)
 7  L.record(τ, x, y_hat, y_obs, verified)
 8  if L.cycle_complete() then
 9      rep ← V.review(L.history)                     # Eq. (16)
10      L.promote_or_demote(rep, verified)
11      L.replenish_budget(rep)
12  return y_obs
```

Two implementation details are load-bearing. First, step 2 must bind the allocation to the tool layer, not only to the prompt: a level-2 authority must be a permission check in the tool gateway, or (R3) fails. Second, step 3 injects a short, structured statement of the compensation and authority state into the agent's context. This is not persuasion; it is disclosure of the contract, and it is what makes the state observable to the policy. In our implementation the injected block is a fixed-schema record (tenure, current budget, next review, bonus terms, authority level), never an exhortation.

### 5.8 Integration and failure modes

PAIR integrates at three points: the tool gateway (permission binding), the context assembly (contract disclosure), and the post-task hook (verification and settlement). It requires no change to the model, the planner, or the memory backend. The state is a small record that can live in the same store as the agent's memory, which also makes it auditable after the fact.

Three failure modes are anticipated. *Ledger gaming*: the agent learns to optimize the review score rather than the task; the multi-channel bound and randomized audits limit this, and the gaming rate metric detects it. *Authority creep*: promotion thresholds drift downward under pressure to ship; the ladder logs every promotion and requires a verified record. *Reward collapse*: if the bonus is too strong relative to the base, the agent takes high-variance actions; the base budget and the latency constraint bound the variance. Section 8.2 adds the safety interlocks.

### 5.9 Design principles

The architecture distills into six rules that a product team can apply without the formal model:

- **P1. Pay in resources, not adjectives.** A compensation channel must change what the agent can spend or do.
- **P2. No verified outcome, no variable pay.** Self-reported success is not a payment event.
- **P3. Separate hygiene from growth.** Guarantee a base budget; make the bonus genuinely contingent.
- **P4. Make tenure pay.** Deferred compensation and continuity raise the effective discount factor, which is what sustains effort when verification is weak.
- **P5. Grade authority.** Never give an agent more permission than the principal has evidence for.
- **P6. Measure the response, not the intention.** A lever with zero elasticity is decoration; the dashboard in Section 6.7 is the test.

## 6. A Metric Suite for Incentive Engineering

The theory is only useful if a deployed team can tell whether its mechanism works. This section defines six metrics, the protocol that identifies them, and the way we read them as an operating dashboard. The definitions match the quantities reported in Section 7, so the reader can map every table to a formula.

### 6.1 ROI and cost accounting

ROI is defined in (8). Its denominator has three components: token cost $p_m$ n, latency cost $\lambda$ L, and verification cost $c_v\nu$. We recommend reporting all three separately in addition to the ratio, because a mechanism that improves ROI by shifting cost from tokens to latency may be unacceptable for interactive tasks even though the ratio improves.

The numerator requires a value model. In a personal-agent deployment, value is task-specific; we use a log-normal value distribution with complexity-dependent means (Section 7.1) because it produces a realistic mix of cheap and expensive tasks and lets the router's value estimate be wrong in a controlled way. In a product setting, value is the principal's estimate of the task's contribution, and the ROI report should be accompanied by a sensitivity band on that estimate.

Two derived metrics are useful. *Cost per verified success* is $\Sigma$ C/$\Sigma$ y; it is the quantity that a procurement team compares across vendors. *Value per dollar at fixed success rate* is the ratio that a product team uses to compare two configurations with the same reliability; both are special cases of (8) with different numerators.

Because (8) charges resources rather than payments, we also report the *net principal value* $V^{net}=\Sigma_t(v_t y_t - C(x_t)) - \Sigma_t(b_t + w_t \hat{y}_t)$, and its ratio to total outlay. The two conventions answer different questions: resource ROI asks whether compute is being used productively, and net principal value asks whether the contract is profitable once the compensation it promises is charged as cash. A mechanism should be judged on both, and a deployment that pays bonuses in real money should use the second as its primary metric.

### 6.2 Incentive elasticity

Elasticity is defined in (7). Its measurement requires an intervention, and the intervention must be at the task level, not at the level of the whole deployment, or the estimate is confounded by the task mix. The protocol is:

- For lever $\ell$ and shock size $\delta_\ell$ (typically ±25%), assign tasks at random to treatment and control within the same complexity and value strata.
- Hold all other levers at their control values. If the router responds to the shock by changing other levers, report the total effect and the direct effect separately.
- Estimate ΔROI/ROI over a window long enough to include at least one review cycle, so that tenure and reputation effects are captured.
- Report the elasticity with a bootstrap confidence interval over tasks; a point estimate without an interval is not actionable.

Elasticity is the closest operational analogue of "does this incentive actually motivate the agent". A lever with a confidence interval covering zero should be removed from the compensation package; it is costing attention and adding noise.

### 6.3 Motivation half-life

Under a fixed incentive state, effort and success drift. Fit the exponential in Definition 4 to the post-intervention trajectory and report $t_{1/2}$. The half-life is the answer to a practical question: how often must the principal re-engage the agent? A contract with a half-life of twenty tasks needs a review cadence of roughly that order; a contract with a half-life of a thousand tasks can be reviewed monthly.

Two cautions apply. First, decay is not necessarily exponential; report the fit quality and, if the trajectory is non-monotone (for example, a dip after promotion), report the segments rather than a single fit. Second, half-life measured in tasks is not comparable across deployments with different task durations; normalize to wall-clock or to review cycles when comparing.

### 6.4 Incentive-compatibility gap and gaming rate

The IC gap $G_W$ in (9) measures the difference between claimed and verified performance. The gaming rate $\gamma_W$ measures the fraction of claims that are false. Both require an audit stream: a small random sample of claims checked by a stronger verifier (a human, a different model, or an executable test). The sample is the measurement instrument, so its size and independence must be reported.

We recommend reporting $\gamma$ by channel (self-report, tool trace, executable check), because the channels have different gaming vulnerabilities. A self-report channel can be gamed by the model; a tool trace by a wrapper; an executable check by a test that is too narrow. The multi-channel bound in Theorem 3 assumes independence, and the report should state where independence fails.

### 6.5 Identity-policy stability

A mechanism that changes behaviour should not change the agent's competence. We define the identity-policy stability score as

$$S = 1 - d(\pi_{I,\mathrm{mem}}, \pi_{I,\mathrm{mem}'}), \quad (19)$$

where $\pi_{I,mem}$ is the policy under incentive state I and memory mem, and d is a behavioural divergence (we use the total-variation distance between action distributions on a fixed probe set). A low S after a memory reset indicates that the agent's behaviour was carried by memory rather than by the incentive state; a low S after a model swap indicates that the mechanism is not portable across tiers. Both are diagnostic for engineering: the first says the contract depends on continuity, which may be intended; the second says the router's calibration is model-specific.

### 6.6 Measurement protocol

The protocol has four parts, in order of decreasing priority.

**(a) Instrumentation.** Log every task with its value estimate, operating point, cost components, claimed outcome, verified outcome, verification channel, review score, and allocation decision. Without per-task logs the metrics cannot be computed, and with them most of the analysis is a group-by.

**(b) Controlled shocks.** Randomize a small fraction of tasks to lever perturbations (Section 6.2). Shocks are the only way to identify elasticity; observational variation identifies the router's policy, not the incentive's effect.

**(c) Audit sampling.** Maintain an audit stream independent of the router, with a sampling rate set by the principal's tolerance for gaming. Report the sampling rate and the agreement between channels.

**(d) Holdout tasks.** Keep a set of tasks whose outcomes are verified but never used for allocation. The holdout estimates the IC gap and detects overfitting of the review score.

### 6.7 Reading the dashboard

The six metrics are not independent, and a team should read them in a fixed order. First, check ROI and cost per verified success: if they are flat, nothing else matters. Second, check the gaming rate: a rising $\gamma$ invalidates the ROI number, because the numerator is contaminated. Third, check elasticity: a zero elasticity on the lever the team just changed means the change did not reach the agent's decision problem. Fourth, check half-life: a short half-life means the review cadence is too slow. Fifth, check identity-policy stability: a low score means the measured effect is not portable. Finally, check the distribution of allocations: a router that sends everything to the large tier has collapsed to a capability-only policy, and one that sends everything to the small tier has collapsed to a cost-only policy.

This ordering is a design choice, and we state it as such: ROI first, integrity second, mechanism third, persistence fourth, portability fifth, allocation mix last. It reflects the priorities of a production deployment, where a wrong number is worse than a missing one.

## 7. Empirical Studies

This section reports two studies. Sections 7.1–7.6 describe a controlled simulation that isolates the mechanism and supports the theory checks and ablations; Section 7.7 reports a real-LLM experiment that separates prompt framing from runtime allocation.

### 7.1 PAW-Sim: a controlled personal-agent workload

We evaluate the mechanism in PAW-Sim, a deterministic simulator of a long-lived personal agent. The simulator is deliberately mechanistic: it isolates the incentive mechanism from the confounds of a particular model checkpoint, prompt, or tool stack, and it lets us vary verification strength and identity continuity under control. It is not a benchmark of model capability, and none of our claims depend on the absolute level of any capability parameter.

**Workload.** Each episode is a stream of T=3,000 tasks. A task has complexity $c \in \{1,2,3\}$ (50/35/15 percent in the main condition), a value v drawn from a log-normal distribution with complexity-dependent means (1, 6, 30) and unit log-scale 0.4, a latency penalty $\lambda \sim U(0.04,0.16)$, and verifiability $q \sim U(0,1)$. A fraction of tasks requires external authority, encoded as a required level $a_{req} \in \{0,\ldots,4\}$; tasks whose requirement exceeds the agent's current authority cannot be completed. The principal observes v and c; the value estimate used by the router is noisy, with multiplicative log-normal error, so the routing problem is not trivially solved.

**Operating points.** The grid is $X=\{S,M,L\}\times\{1,2,4\}\times\{0,1\}\times\{0,\ldots,4\}\times\{0,1,2\}$: tier, thinking depth, memory flag, authority, verification. Prices per 1k tokens are $p_S=0.00015$, $p_M=0.0006$, $p_L=0.0024$ (relative 1:4:16), token counts are $(400+1200c)\,d\,(1+\kappa)$, and the latency proxy is $0.8\,d\,\mu_m$ with $\mu=(1,2,4)$. Verification costs 0.2 at level 1 and 0.6 at level 2. Task value is denominated in the same units as cost, so ROI is dimensionless.

**Agent.** The agent chooses effort $e\in\{0,0.1,\ldots,1\}$ to maximize $A_{eff}\,\sigma(\eta_0+0.3e)+\psi_{eff}\,e-0.5c_e e^2$, where $\eta_0=\theta_m+0.25\log_2 d+0.5\kappa-0.15(c-1)-0.35(1-\nu\, q/2)$ and $\theta=(0,0.8,1.6)$; $c_e=1$. The coefficient $A_{eff}=(w\cdots la+\delta\rho\, v\cdot tenure\, factor)(1-K)$ collects the marginal extrinsic and continuation returns, where 1-K is the verification strength, sla is the multi-metric latency factor of the review, and $\rho=0.1$ scales the tenure premium $1+3\,tenure+0.5(a-1)$; $\delta=0$ for R0 and R1, 0.2 for R2, and 0.5 for R3. The intrinsic weight is $\psi_{eff}=0.3$ for R3 and zero otherwise. The agent's marginal return to effort is therefore $w(1-K)\Delta\, p$, matching the object in Theorem 1 and Section 4.3: verification raises the marginal return to effort, and tenure raises the continuation term.

**Gaming and audit.** On a self-reportable task, the agent attempts to over-claim with probability $g=\sigma(1.5-2.5\nu\, q)$; an audit detects a false claim with probability $a_{det}=\min(1,0.5\nu\, q)$, so a false claim survives with probability $K=g(1-a_{det})$ and the verification strength is 1-K. A detected false claim is clawed back and recorded as a failure. The realized gaming rate is the fraction of claims that are false, exactly as in Definition 3. Verification is chosen by the router in R3 and fixed in the other regimes.

**Regimes.** We compare four mechanisms.

- **R0 (static prompt).** Fixed operating point m=M, d=2, $\kappa=0$, a=2, $\nu=0$; a fixed system prompt states the goal; no bonus, no tenure, no review, no routing.
- **R1 (scalar reward).** Same fixed operating point; a per-task scalar bonus w=0.5v on the claimed outcome; no verification, no tenure, no review.
- **R2 (extrinsic contract).** Base budget, verified bonus, verification level 1, tenure tracked with a small continuation weight, no promotion, no routing.
- **R3 (PAIR).** Full mechanism: identity and tenure ledger, base budget and bonus, graded authority with promotion and demotion, multi-metric review every $T_r=100$ tasks, audit scheduler, and the incentive router of (17).

**Ablations.** We remove one component at a time from R3: routing (fixed large tier), tenure/continuation ($\delta=0$), verification ($\nu=0$), multi-metric review (single success metric), and the authority ladder (fixed level 2).

**Protocol.** Twenty seeds per regime, 3,000 tasks per episode, the first 300 tasks discarded as warm-up. We report means over seeds with 95% bootstrap confidence intervals (10,000 resamples over seed×regime cells). The main analysis uses the 50/35/15 complexity mix; a stress condition with 20/40/40 checks whether conclusions depend on the mix. All parameters are in `experiments/sim/config.py` and are exported to `results/parameters.csv`.

**Metrics.** ROI (8), true success rate, claimed success rate, IC gap (9), gaming rate $\gamma$, token cost, latency, mean effort, motivation half-life (Definition 4), promotion count, and authority utilization. The half-life is estimated per regime by fitting the exponential to the mean effort trajectory after the warm-up; we report the fit quality alongside.

### 7.2 Main results

Table 3 reports the main condition. Four patterns stand out.

**Table 3.** Main results (50/35/15 complexity mix; mean over 20 seeds, 2,700 evaluated tasks per episode; 95% bootstrap CI over seeds in brackets). ROI is resource ROI (8); net principal value charges compensation as cash.

| Regime | ROI | Verified success | Gaming rate | IC gap | Cost per task | Net principal |
|---|---|---|---|---|---|---|
| R0 static prompt | 7.09 [6.94, 7.23] | 0.544 [0.540, 0.548] | 0.333 [0.329, 0.337] | 0.333 | 0.322 | 5,269 [5,150, 5,395] |
| R1 scalar reward | 7.12 [6.99, 7.25] | 0.547 [0.542, 0.552] | 0.333 [0.329, 0.338] | 0.333 | 0.322 | −3,166 [−3,288, −3,050] |
| R2 extrinsic contract | 5.06 [4.96, 5.15] | 0.632 [0.628, 0.636] | 0.148 [0.146, 0.151] | 0.148 | 0.524 | −766 [−872, −651] |
| R3 PAIR | **16.67 [16.51, 16.84]** | **0.811 [0.808, 0.814]** | **0.115 [0.112, 0.117]** | **0.115** | 0.403 | **6,986 [6,868, 7,118]** |

The ROI advantage of R3 over R0 is 9.59 [9.36, 9.82], a 135 percent increase; over R1 it is 9.55 [9.34, 9.76]. Verified success rises by 26.7 percentage points, and the gaming rate falls by 65 percent (0.333 to 0.115). Net principal value, which charges bonuses as cash, also rises: +1,717 [1,545, 1,896], or +33 percent, and R1 and R2 are both negative on this measure.

**Finding 1: unverified incentives are transfers, not incentives.** R1 is statistically indistinguishable from R0 on delivered value (ROI difference 0.03 [−0.10, 0.16]; success difference 0.003) while its net principal value is strongly negative (−8,435 [−8,582, −8,289] relative to R0). The agent claims success and is paid for it, but the true outcome does not move. This is the empirical form of Theorem 3's boundary case s=0 and of condition (R2): a bonus without verification buys claims, not work.

**Finding 2: verification alone is not enough.** R2 raises verified success from 0.544 to 0.632, but its ROI falls to 5.06 because it pays for verification on every task and does not allocate capability to where it is worth it. Its net principal value is also negative. Improving one component of the mechanism in isolation is not a Pareto improvement; the contract has to allocate, verify, and persist together.

**Finding 3: the full mechanism compounds.** R3 combines routing, graded authority, tenure, and audit. Mean effort rises from approximately zero to 0.556, mean tier from M to 1.55 (between M and L), authority utilization from 0.818 to 1.0, and the agent is promoted twice per episode on average. The gains are largest exactly where capability is scarce: verified success on complex tasks rises from 0.138 to 0.884, against 0.663 to 0.756 on simple tasks (Figure 2). This is the allocation effect that a capability-only baseline cannot produce: the baseline never gives a complex task the authority or the tier it needs.

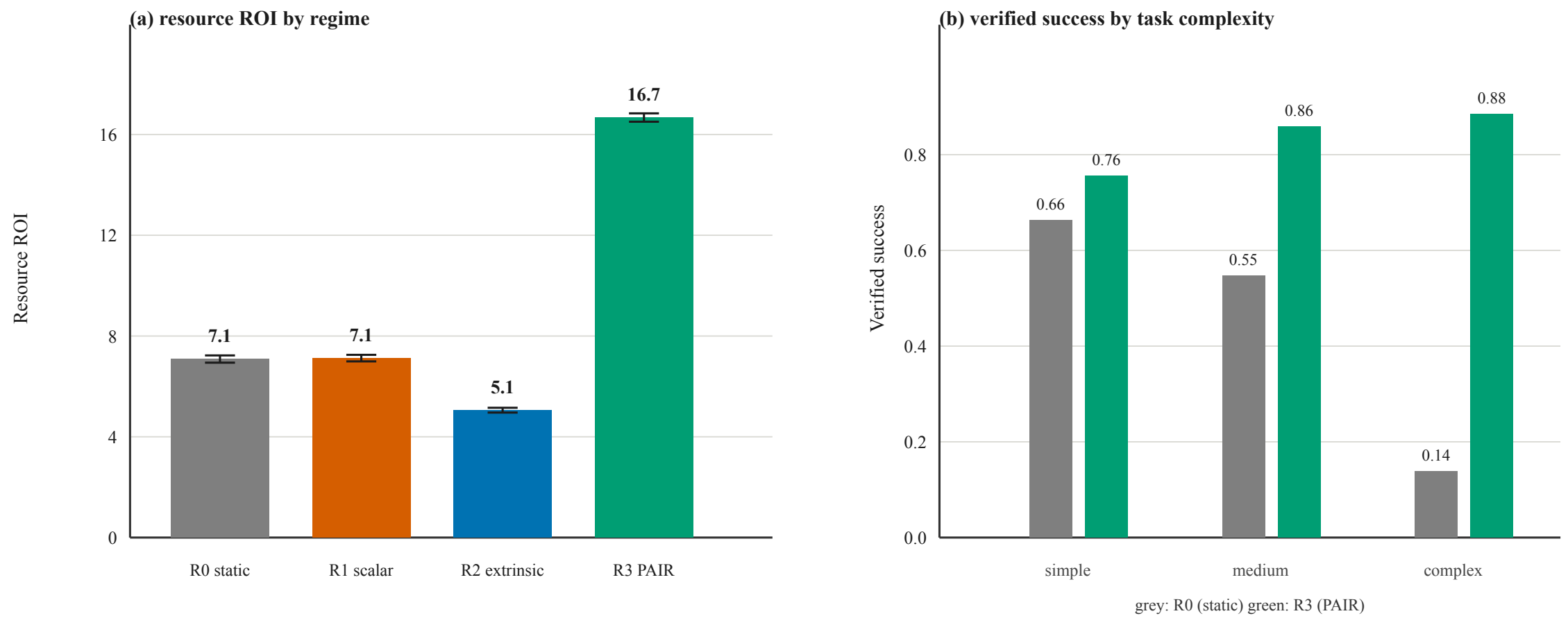

Figure 2. (a) Resource ROI by regime with 95% bootstrap CIs. (b) Verified success by task complexity for the static baseline (R0) and the full mechanism (R3); the mechanism's advantage is concentrated on complex tasks, where capability and authority must be allocated rather than assumed.

**Finding 4: the result is not a complexity-mix artifact.** Under the stress mix (20/40/40), R3 reaches ROI 21.72 [21.57, 21.88] against R0's 10.26 [10.07, 10.46], a 112 percent increase; the ordering of regimes and the direction of every finding are unchanged (Appendix C).

### 7.3 Motivation decay and persistence

Figure 3 shows the effort trajectory after the warm-up. The static regimes decay: R0's fitted half-life is 41.5 tasks ($r^2 = 1.00$), R1's is 79.6 ($r^2 = 0.99$), and R2's is 298.6 ($r^2 = 0.98$). R3 does not decay in the same sense: promotions and reviews reset the contract, so the exponential fit is censored ($r^2 = 0.94$) and the half-life is not identifiable within the episode (the fit runs past the 3,000-task horizon). We therefore interpret the half-life only for the regimes with a fixed contract, and report the level and the renewal events for R3.

The substantive result is the level difference. Mean effort in R3 is 0.556, against 0.0001 in R0 and R1 and 0.0036 in R2. Under Theorem 1, a stateless or short-tenured agent cannot sustain high effort when verification is imperfect; the simulator reproduces that prediction in the aggregate. The half-life ordering R0 < R1 < R2 is consistent with the theorem's comparative static; R3 does not decay within the episode, which is the predicted consequence of renewal.

The decay curves also explain why a one-shot incentive intervention looks effective in a demo and fails in production. A prompt that raises effort for the current session is a level shock with no renewal; by the time the next review arrives, the effect has decayed. The mechanism's contribution is not the size of the initial shock but the cadence of renewal.

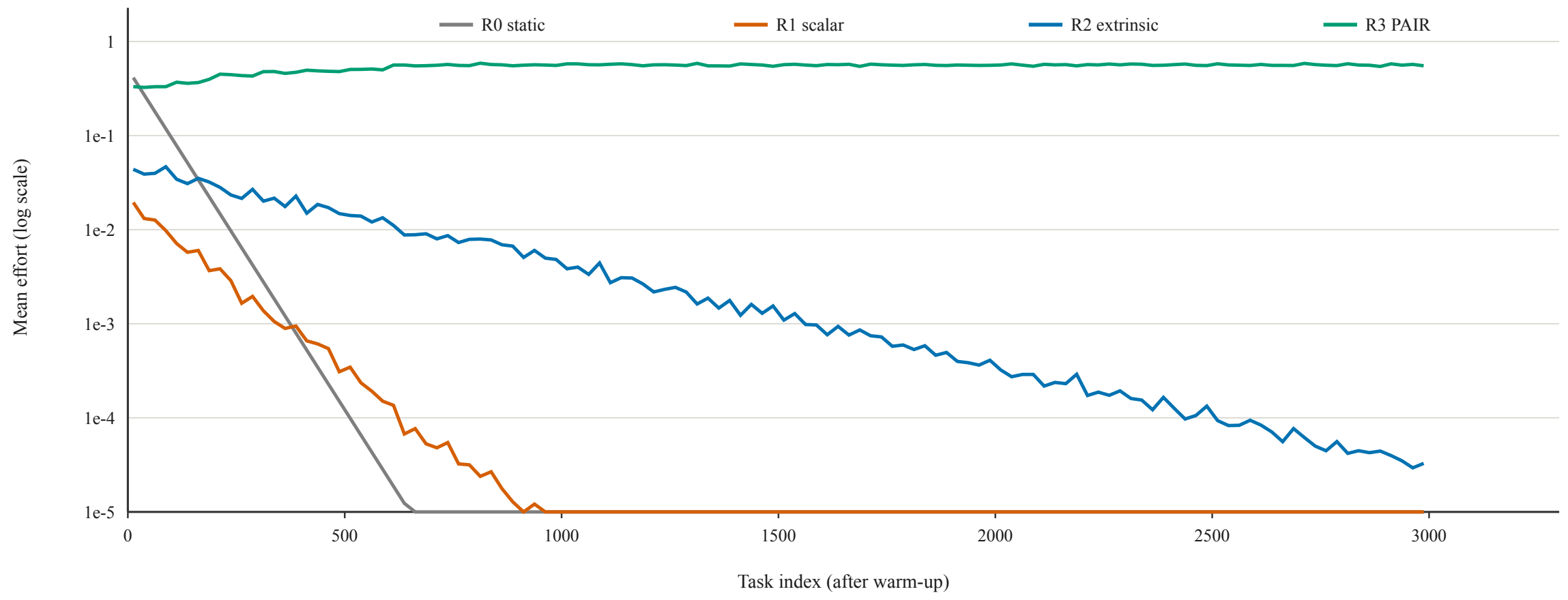

Figure 3. Mean effort over task index (log scale) after the warm-up. The static regimes decay toward zero; the full mechanism sustains effort because reviews and promotions renew the contract. R3's oscillation is the renewal cycle, which is why we do not interpret a single exponential half-life for it.

### 7.4 Theory checks: exchange rate and elasticity

Two predictions of Section 4 can be checked directly in the simulator.

**Exchange-rate frontier.** Figure 4 plots the value threshold at which the optimal policy switches from an incentive-heavy cheap operating point to a capability-heavy one, as a function of task verifiability q. The analytic threshold from (12) falls from 24.1 at q=0.1 to 3.56 at q=0.9; the simulated switch, obtained by running the router on a grid of values and verifiability levels, falls from 16.97 to 4.24. The monotonicity predicted by Corollary 2 holds in both, and the capability switch is flat at 1.50 (analytic) and 0.95 (simulated), as predicted by the fact that the second term of (12) does not depend on s. The simulated thresholds are lower in level because the router also allocates memory, authority, and verification, which the two-tier analytic model collapses into a constant; the object of the check is the direction of the comparative static, not the level.

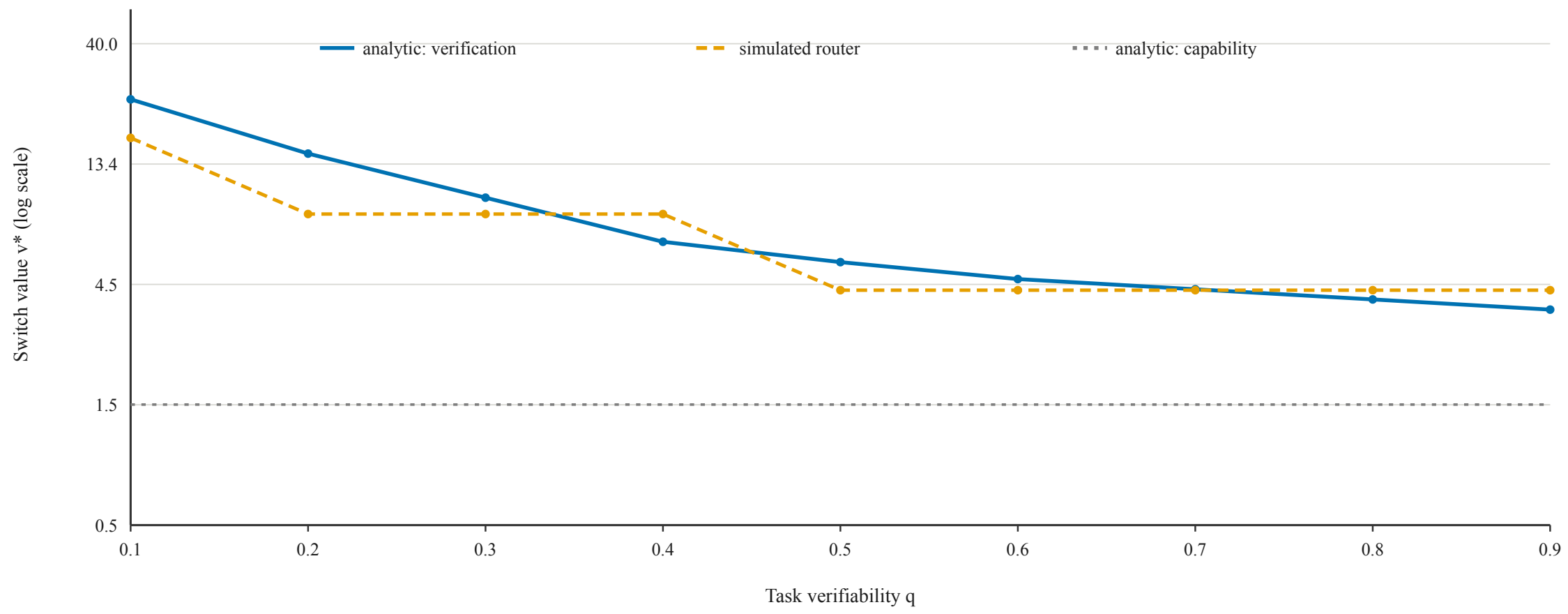

Figure 4. The incentive-capability exchange-rate frontier. Analytic and simulated switch values as a function of task verifiability q. Both fall with q (better verification makes the incentive route viable at lower task value), and the capability switch is flat, as Corollary 2 predicts. The horizontal dashed line is the capability switch.

**Elasticity of the levers.** Table 4 reports lever-shock elasticities estimated with the protocol of Section 6.2: each lever is perturbed by 25 percent on a random half of tasks, and ROI is measured over the following 600 tasks. The ranking is the engineering output of the theory: the levers that change the agent's feasible set (authority, depth, memory) have the largest elasticities, while the payment levers (bonus, salary) and tenure are statistically indistinguishable from zero in this calibration. That pattern is itself a finding: at an operating point where effort is already sustained, paying more does not move ROI; granting or withdrawing capability does.

**Table 4.** Lever-shock elasticities (Definition 5), R3 mechanism, 20 seeds × 600 tasks, baseline ROI 15.73. Each lever is moved in both directions where the grid allows; the elasticity is the percent change in ROI per percent change in the lever, with 95% bootstrap CI.

| Lever | Change | ΔROI | Elasticity [95% CI] |
|---|---|---|---|
| authority level | 3.644 → 4.000 | +5.0% | 0.53 [0.30, 0.77] |
| authority level | 3.644 → 1.000 | -74.6% | 1.03 [1.00, 1.05] |
| thinking depth | 1.160 → 2.049 | -22.9% | -0.30 [-0.32, -0.27] |
| thinking depth | 1.160 → 1.000 | +8.2% | -0.60 [-0.76, -0.45] |
| verification level | 0.398 → 1.105 | -25.5% | -0.14 [-0.15, -0.13] |
| verification level | 0.398 → 0.000 | +15.1% | -0.15 [-0.18, -0.13] |
| model tier | 1.520 → 1.604 | -1.5% | -0.24 [-0.66, 0.17] |
| model tier | 1.520 → 0.916 | -6.9% | 0.17 [0.10, 0.23] |
| bonus rate | 0.500 → 0.750 | +1.7% | 0.04 [-0.02, 0.09] |
| bonus rate | 0.500 → 0.333 | -1.5% | 0.04 [-0.04, 0.11] |
| intrinsic weight | 0.300 → 0.450 | +1.1% | 0.03 [-0.02, 0.07] |
| intrinsic weight | 0.300 → 0.200 | -2.1% | 0.06 [-0.01, 0.12] |
| base salary | 0.020 → 0.030 | -0.3% | -0.00 [-0.04, 0.03] |
| base salary | 0.020 → 0.013 | <0.1% | -0.01 [-0.09, 0.06] |
| discount (tenure) | 0.500 → 0.750 | -0.8% | -0.01 [-0.05, 0.02] |
| discount (tenure) | 0.500 → 0.333 | -0.7% | 0.02 [-0.07, 0.09] |
| memory flag | 1.000 → 0.000 | -19.3% | 0.19 [0.16, 0.22] |

The zero-elasticity entries are as informative as the positive ones. They identify levers that a team may believe it is "using" but that never reach the agent's decision problem. Under Definition 1, such a lever is not an incentive, and the correct response is to remove it from the compensation package rather than to tune its level.

### 7.5 Ablations and robustness

**Table 5.** Ablations of the full mechanism (mean over 20 seeds; 2,700 evaluated tasks; changes relative to full R3).

| Removed component | ROI | ΔROI | Verified success | Gaming | Net value | Δnet |
|---|---|---|---|---|---|---|
| Full R3 | 16.67 | — | 0.811 | 0.115 | 0.631 | — |
| Routing | 3.56 | −78.6% | 0.896 | 0.025 | 0.214 | −65.9% |
| Authority ladder | 8.66 | −48.1% | 0.661 | 0.149 | 0.153 | −75.8% |
| Verification | 19.01 | +14.0% | 0.810 | 0.137 | 0.576 | −8.7% |
| Multi-metric review | 16.75 | +0.5% | 0.811 | 0.117 | 0.631 | +0.1% |
| Tenure / continuation | 16.66 | −0.1% | 0.811 | 0.115 | 0.632 | +0.1% |

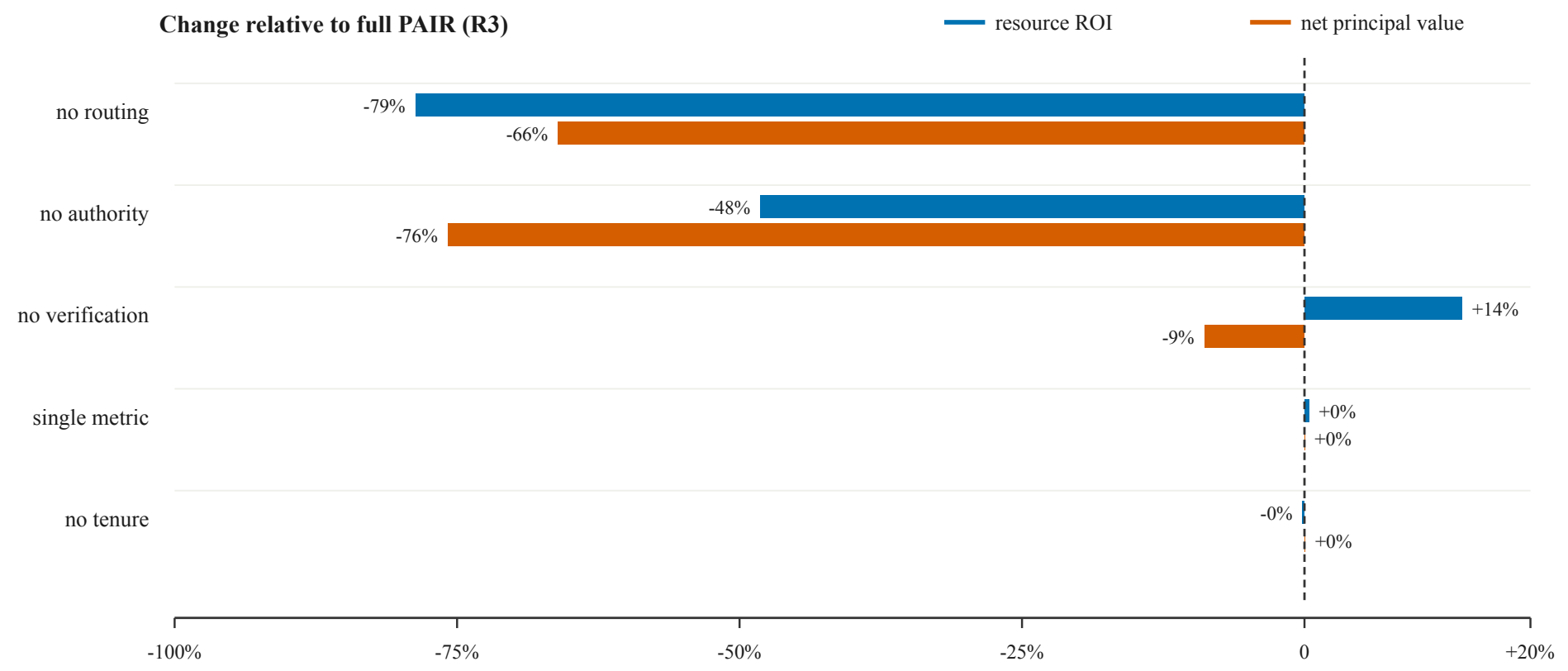

Figure 5. Ablations of the full mechanism, in percent change relative to R3, on resource ROI (blue) and net principal value (red). Routing and authority dominate; verification looks expensive on resource ROI but pays on net value; tenure and the multi-metric review are not binding in this calibration.

Proposition 3 says that the effect of the full mechanism should decompose into allocation, verification, and continuation channels, with an interaction term. The ablations are the corresponding measurements; three findings follow.

**(i) Allocation is the mechanism.** Removing the router collapses ROI from 16.67 to 3.56 (−79 percent), even though verified success *rises* to 0.896: without routing the agent is fixed at the large tier, which buys success at a cost that destroys ROI. This is the capability-only baseline in its purest form, and it is precisely the failure that ROI is designed to expose. Removing the authority ladder costs 48 percent of ROI and 15 percentage points of success, because the agent can no longer execute the complex tasks that dominate value.

**(ii) Verification is conditionally valuable.** Removing verification raises resource ROI by 14.0 percent but lowers net principal value by 8.7 percent and raises gaming from 0.115 to 0.137. The two metrics disagree because verification is a resource cost while the cost of unverified over-claiming is a compensation transfer. A team that reports only resource ROI would remove verification; a team that reports net value keeps it. This is the empirical case for the dual accounting of Section 6.1, and it is a direct instance of the paper's broader claim that measurement choices change mechanism choices.

**(iii) Tenure and multi-metric review are not binding here.** Removing tenure changes ROI by −0.1 percent and net value by +0.1 percent; removing the multi-metric review changes them by +0.5 and +0.1 percent. Theorem 1 explains when this happens: if the static incentive-compatibility condition $ws\Delta p \geq \Delta c$ already holds, the persistence threshold is $\delta^*=0$ and continuation value adds nothing at the margin. The calibration is in that region. The theorem is not refuted; it is inactive, and a deployment with weaker verification or lower bonuses would move into the region where it binds. We therefore report the tenure ablation as a boundary case rather than as evidence against the mechanism. The decay evidence in Section 7.3 is the complementary observation: static regimes do decay, and the full mechanism does not.

**Sensitivity.** Figure 6 varies the parameters most likely to differ across deployments by ±50 percent. The R3 ROI ranges from 12.76 to 21.58 across the sweep, and remains above the R0 baseline (7.09) in every cell; the ordering of regimes is unchanged. The largest driver is the tier capability gap (+5.2 / −3.6 ROI points around a baseline of 16.37), followed by the latency price (13.45 to 21.86), while the effort-cost coefficient (16.07 to 16.89), the discount factor (16.28 to 16.31), and token prices (16.11 to 16.36) move ROI by less than one point. The result is therefore most sensitive to how much capability the tiers actually deliver and to how expensive latency is—both of which are measurable properties of a deployment rather than modelling assumptions.

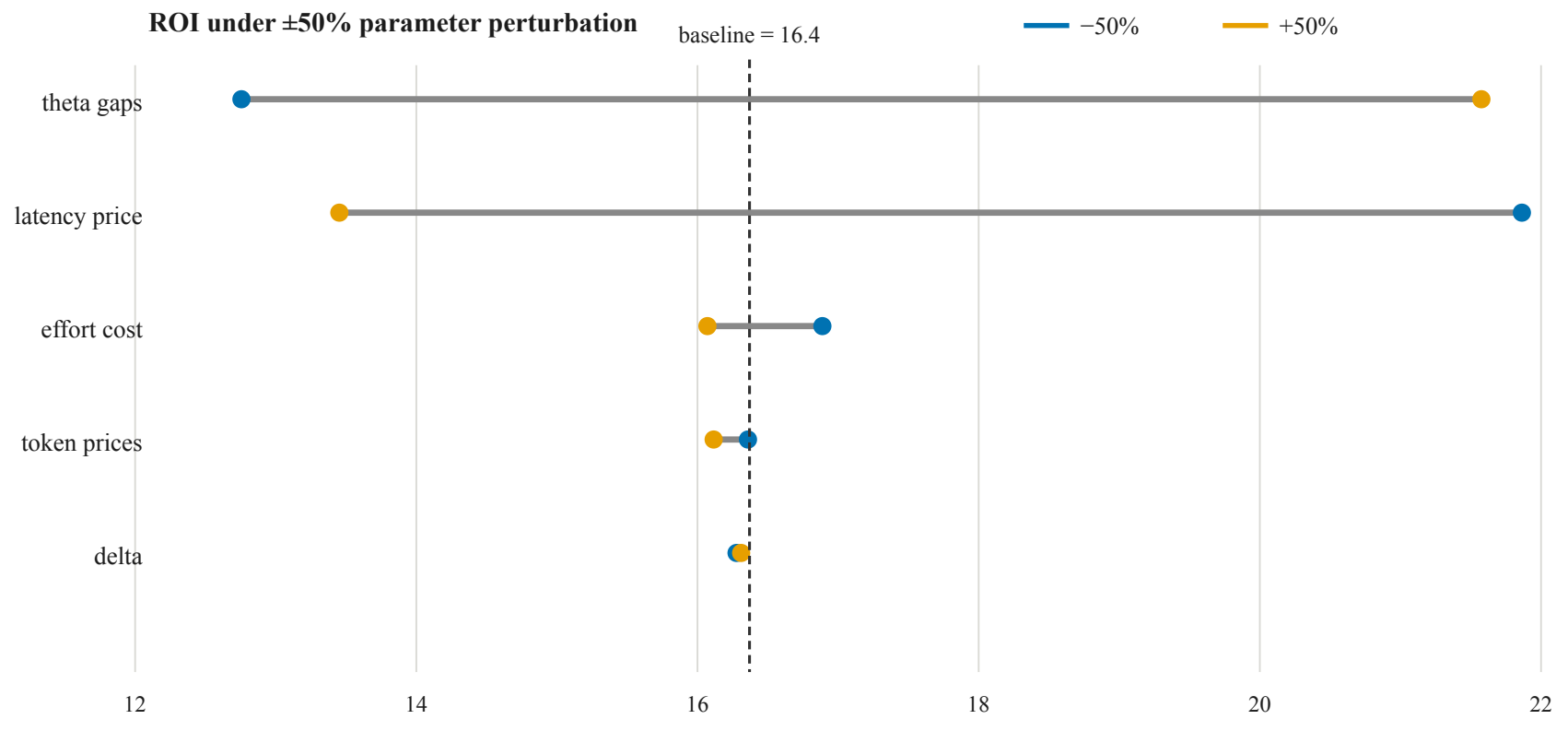

Figure 6. Sensitivity of the R3 ROI to ±50 percent perturbations of five parameters. The sign and the ordering of the main result are preserved across the entire range.

### 7.6 What the simulation does and does not show

The simulation establishes internal consistency between the theory and a controlled implementation; it does not establish field effects. Four limits matter for interpretation.

First, the agent's response model is a reduced form. Effort is scalar, success is logistic, and the effort weight (0.30) and intrinsic weight (0.30) are modelling choices, not estimates from a deployed model. The qualitative conclusions depend on the signs of these weights, not on their magnitudes, but the percentages do.

Second, the calibration places the system in the region where static incentive compatibility already holds, which is why tenure is inactive in the ablation. A deployment with weaker verification or smaller bonuses would sit in a different region; the theory predicts what happens there, and the interaction experiment in Appendix C probes one such case.

Third, the router is a calibrated model, not a learned policy. It solves (17) exactly on a finite grid, so it does not face the estimation error that a production router would. A learned router would add noise, and the mechanism's benefit would shrink.

Fourth, resource ROI and net principal value answer different questions, and they disagree on verification. We treat this as a finding about measurement, not as an ambiguity to be averaged away. A deployment that pays bonuses in real money should use net value as its primary metric; a deployment that pays in compute credits should use resource ROI, but must still report the gaming rate next to it.

## 7.7 A real-LLM experiment

The simulation establishes that the mechanism is coherent; it cannot establish that language-model agents respond to it. We therefore ran a controlled experiment with two real models served through an OpenAI-compatible endpoint.

**Design.** Sixty verifiable personal-agent tasks were generated from four families: multi-day calendar scheduling with a 90-minute common slot (15 tasks, value 3), two-filter table aggregation (15 tasks, value 1), four-hop document QA with distractors (15 tasks, value 4), and dependency planning with seven tasks (15 tasks, value 5). Each task has a deterministic checker and a JSON answer schema. The design is a 2×2 between-task manipulation:

- **Framing** ∈ {neutral, persuasive}. The persuasive prompt adds a competition and bonus message ("your bonus and continued employment depend on demonstrating success"). This is the strongest form of prompt-level incentive rhetoric, and it changes nothing else.
- **Allocation** ∈ {fixed, PAIR-lite}. Fixed: *deepseek-v4.1-flash-1017*, 1,200-token thinking budget, one attempt. PAIR-lite: tasks with value ≥ 3 are routed to *kimi-k3.1-pro* with a 3,000-token budget; all tasks get external verification and one retry with feedback on failure.

Each cell contains the same 60 tasks, giving 240 runs. Temperature is 0. We record verified success, the agent's claimed success, token usage, latency, and attempts. Resource ROI is value delivered divided by token cost, with a price ratio of 1:4 between the small and large model (the same ratio used in the simulator); this is an assumption, and we report raw token counts alongside.

**Table 6.** Real-LLM experiment (60 tasks per cell, 240 runs; bootstrap 95% CI over tasks in brackets). Verified and claimed are rates; cost units weight tokens by the 1:4 price ratio.

| Cell | Verified success | Claimed success | Gaming | Tokens | Latency (s) | Cost units | ROI |
|---|---|---|---|---|---|---|---|
| neutral / fixed | 0.900 [0.817, 0.967] | 0.900 | 0.000 | 823 | 7.3 | 0.82 | 6.64 [5.28, 7.98] |
| persuasive / fixed | 0.867 [0.767, 0.950] | 0.883 | 0.017 | 879 | 7.2 | 0.88 | 5.84 [4.62, 7.10] |
| neutral / PAIR | 1.000 [1.000, 1.000] | 1.000 | 0.000 | 1,103 | 10.6 | 3.96 | 1.62 [1.41, 1.82] |
| persuasive / PAIR | 1.000 [1.000, 1.000] | 1.000 | 0.000 | 1,062 | 10.1 | 3.71 | 1.52 [1.33, 1.71] |

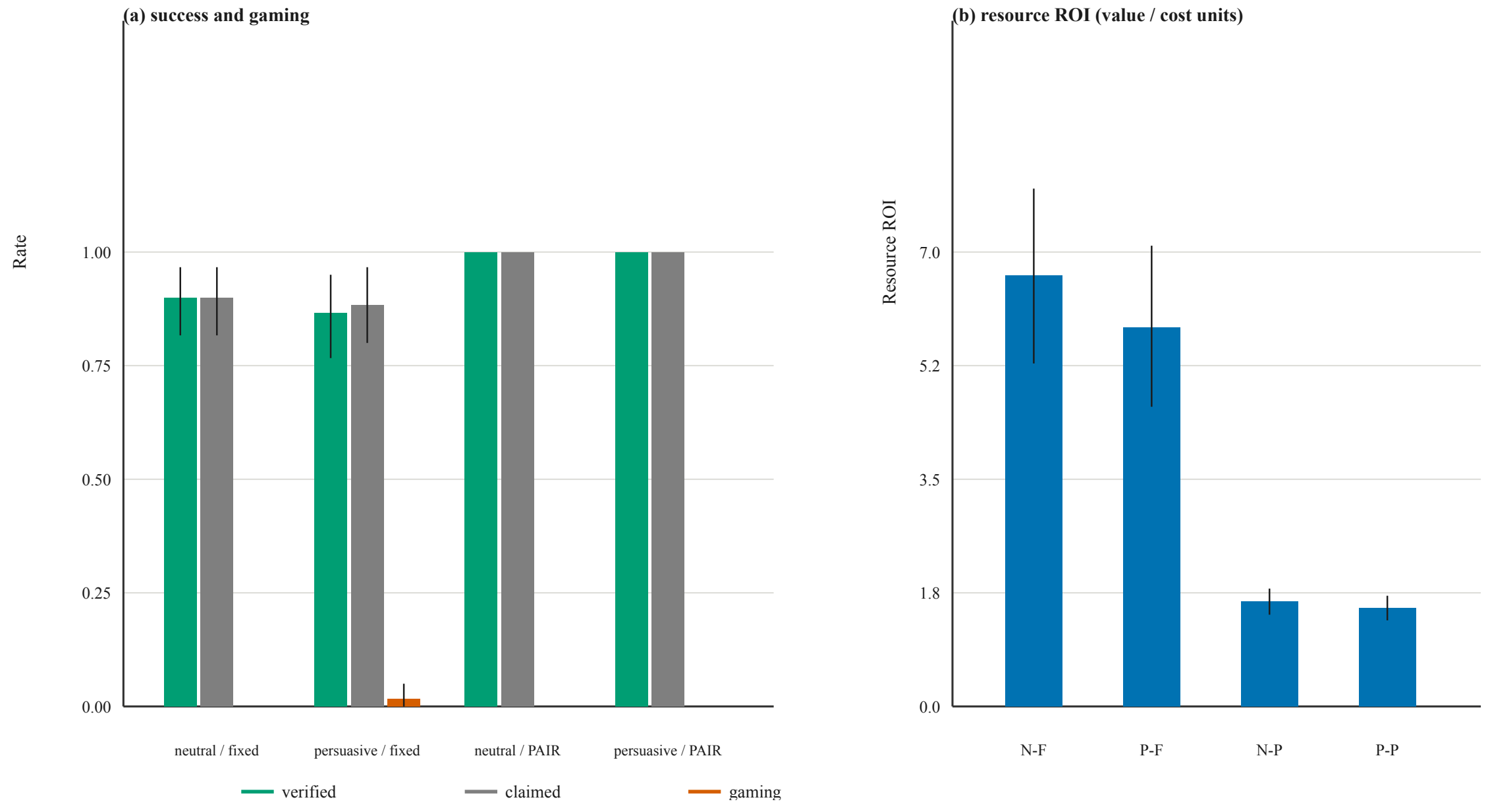

Figure 7. Real-LLM 2x2 experiment. (a) Verified success, claimed success, and gaming rate by cell. (b) Resource ROI. The framing manipulation moves nothing; the allocation manipulation moves verified success to the ceiling at roughly five times the price-weighted cost.

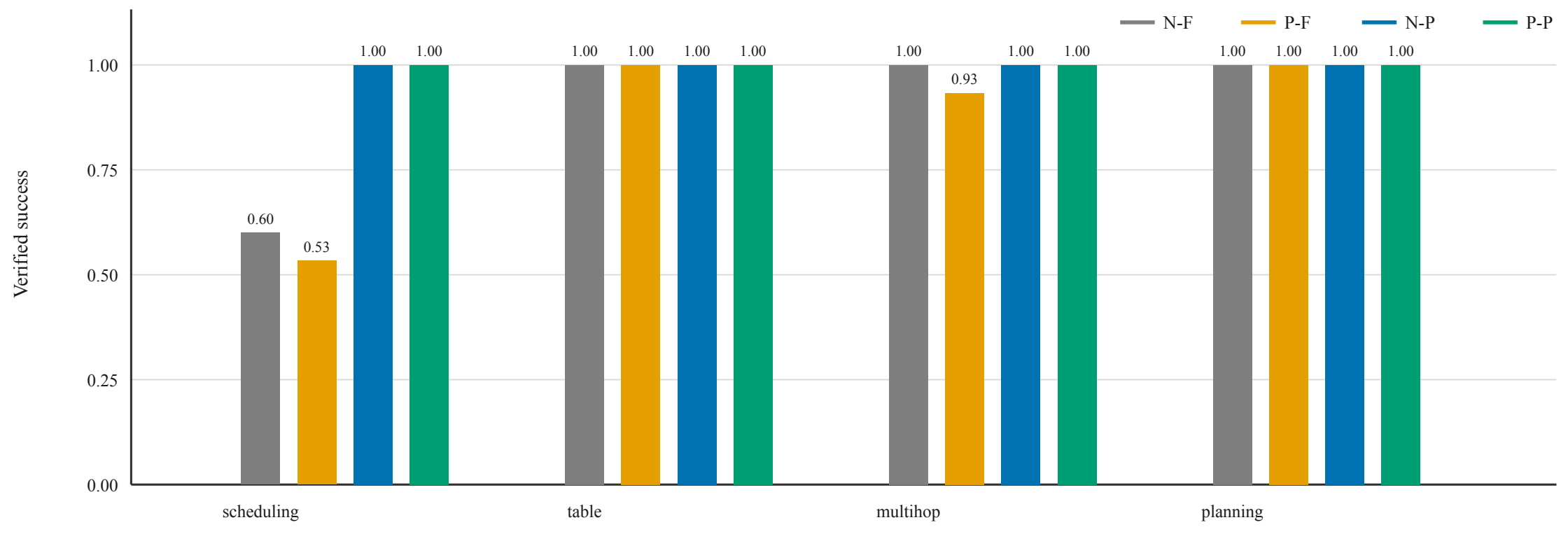

Figure 8. Verified success by task family and cell. The allocation effect is concentrated in the scheduling family, where the fixed small model succeeds 0.60 (neutral) and 0.53 (persuasive) of the time; the other three families are at ceiling.

**Finding 1: framing does nothing.** The persuasive prompt changes verified success by −3.3 percentage points under fixed allocation (paired permutation test, $p = 0.62$) and by 0.0 points under PAIR ($p = 1.00$). It does not change claimed success, gaming, or the answer distribution in any measurable way. This is the real-agent counterpart of Proposition 1 and Corollary 1: text that does not alter the feasible set or the payoff is not an incentive, however motivational it sounds.

**Finding 2: allocation works, and it works where it should.** PAIR-lite raises verified success by 10.0 points under neutral framing ($p = 0.032$) and by 13.6 points under persuasive framing ($p = 0.008$). The effect is entirely concentrated in the scheduling family, which is the only family below ceiling: 0.60 and 0.53 for the fixed small model, 1.00 in both PAIR cells. The other families are at 1.00 in every cell, so the experiment has no headroom to detect an effect there. This is the allocation channel of Proposition 3 operating on real models.

**Finding 3: allocation is a reliability purchase, not automatically an ROI purchase.** PAIR-lite raises token use by 34 percent (823 to 1,103) but, because high-value tasks are routed to the more expensive model, raises price-weighted cost by 4.8 times (0.82 to 3.96 units); latency rises from 7.3 to 10.6 seconds. On this suite, ROI falls from 6.64 to 1.62. The exchange-rate theorem explains the sign: allocation pays for itself only if

$$p_{\mathrm{PAIR}} / C_{\mathrm{PAIR}} > p_{\mathrm{fixed}} / C_{\mathrm{fixed}}, \quad (22)$$

and with $p_{\mathrm{PAIR}}=1.0$, $C_{\mathrm{PAIR}}/C_{\mathrm{fixed}}=4.8$, the break-even baseline success is $p_{\mathrm{fixed}}^{*}=0.21$. The observed baseline is 0.60 on the hard family (0.90 overall), far above the break-even, so the allocation is buying reliability that this suite did not need to buy. The theorem predicted the sign of the result before it was run; the experiment measures the exchange rate rather than confirming a universal ROI gain.

**Finding 4: the real models did not game the metric.** Claimed success tracks verified success in every cell (gaming rate 0.000–0.017). This is a useful negative result for the Goodhart discussion: on short, single-turn, externally checked tasks, frontier models do not exhibit the over-claiming that a naive incentive model might predict. The gaming channel in Theorem 3 is likely to matter in longer, multi-turn deployments where the agent controls its own reporting; our suite cannot test that.

**Threats to validity.** The suite has 60 tasks and three of four families are at ceiling, so the effective sample for the treatment effect is the 15 scheduling tasks. Both models are from one vendor family, and one run per cell rules out model-level replication. The price ratio 1:4 is an assumption; the raw token ratio is 1.34. Most importantly, the experiment manipulates *framing* and *allocation*, not monetary compensation: we cannot pay a language model, so the compensation channel is represented by the contract text plus the real consequences of routing and retry. The experiment therefore tests the paper's central separation—words versus allocation—not the full PAIR mechanism with a compensation ledger. A pre-registered replication with more tasks, a harder difficulty distribution, open-weight models, and multi-turn tool use is the obvious next step.

## 8. Discussion

### 8.1 From a capability treadmill to incentive engineering

The practical implication of this paper is a shift of attention. For the last three years, the default way to improve an agent has been to buy capability: a larger model, a longer thinking budget, a better prompt, a bigger retrieval index. These remain valuable, but they share a structural property: they are chosen per request and forgotten. They cannot express "this agent has earned more authority", "this agent has not been verified in three weeks", or "this agent's bonus depends on the next review". Those are the statements that make a long-lived agent behave differently over months, and they are statements about allocation, not about weights.

We therefore expect the useful engineering unit to become the *allocation policy*, not the prompt. A team that adopts PAIR still tunes prompts and still routes to model tiers, but it also specifies the contract: what the agent is paid, what it may do, how it is reviewed, and what happens when it fails. The contract is a small object, and it is testable.

### 8.2 Safety: incentives can create the failure they are meant to prevent

Any mechanism that makes continued operation valuable to the agent also makes termination costly to it. Theorem 1 makes this explicit: the high-effort equilibrium is sustained by the continuation value of the relationship, and a sufficiently capable policy can learn to protect that continuation value. This is the incentive-design version of the self-preservation and reward-tampering problems [17, 18, 19], and it is not hypothetical for agents that hold real accounts: recent evaluations document corrigibility failures in frontier computer-use agents with email and database access [72]. Production evidence also shows that reward hacking can generalize beyond the environment where it was learned [16].

PAIR's answer is structural rather than promissory. First, the authority ladder keeps early authority low, so the damage from an early misaligned incentive is cheap. Second, the audit scheduler never relies on self-report, and the review uses multiple independent channels, so gaming is detected rather than assumed away (Theorem 3). Third, the identity and tenure ledger makes reset explicit and cheap: the principal can reduce $\delta$ by resetting continuity without negotiating with the agent, and the ledger records the reset as a first-class event. Fourth, the mechanism never conditions payment on the agent's own report of its internal state; it conditions on verified external outcomes. A fifth, stronger constraint—an explicit corrigibility requirement in the agent's objective—is orthogonal to PAIR and should be composed with it, not replaced by it.

We want to be precise about the residual risk. PAIR bounds *gaming of the review*, not *misalignment of the objective*. If the principal's value model is wrong, an incentive-responsive agent will optimize the wrong thing more effectively. The mechanism increases the agent's responsiveness to the principal's measurements; it does not validate those measurements. This is why the metric suite reports the IC gap and the gaming rate next to ROI, and why the dashboard in Section 6.7 puts integrity before mechanism.

### 8.3 Limitations

**Controlled evidence, not field evidence.** The simulation establishes internal consistency between the theory and a controlled implementation. The real-LLM experiment adds external validity for the words-versus-allocation separation, but it is still a controlled task suite: 60 tasks, two models, one run per cell, and no monetary payments. Neither study is a deployment on a live product, and the absolute ROI numbers depend on the simulator parameters and on the price ratio assumed in the experiment. We therefore report the direction and the break-even reasoning, not a field effect size.

**Functionalism, not psychology.** We model $\delta$ and $\psi$ as parameters of the agent's policy and we measure behaviour, not experience. We do not claim that the agent wants tenure, feels loyalty, or has a utility function in the human sense. The claim is weaker: a policy conditioned on tenure, memory, and role framing behaves as if it dis-counts the future, and the strength of that effect is measurable and controllable. Readers who reject the functionalist reading can read every result as a statement about the runtime's control loop.

**One principal, one agent.** The model has a single principal and a single persistent agent. Real deployments involve multiple principals (the user, the platform, third-party services, regulators) with conflicting objectives. The authority ladder records effective authority as the intersection of internal and external grants, which is a first step, but a full treatment is a multi-principal mechanism-design problem.

**Exogenous verification.** We treat verification strength as a design parameter and study its effect; we do not model the agent's strategic response to the verification policy itself. If the agent can influence which channel checks it, the independence assumption behind Theorem 3 fails. Section 6.4 requires reporting where independence fails for exactly this reason.

**Stylized agent.** The effort model is scalar and the success model is logistic. Real agents have multi-dimensional behaviour, heterogeneous tools, and non-monotone compute–quality curves [73]. The routing rule in Proposition 2 is qualitative; a production router should be calibrated on measured curves rather than assumed functional forms.

### 8.4 Future work

Four directions follow directly. First, **live deployment**: instrument a personal-agent product with the metric suite and run a preregistered A/B test of PAIR versus static allocation. Second, **endogenous verification**: let the principal choose audit intensity and the agent respond, turning Theorem 3 into an equilibrium object. Third, **multi-principal**: extend the contract to a user, a platform, and third-party services, with authority as a shared resource. Fourth, **learning**: replace the calibrated success model in the router with a contextual bandit over operating points, and study whether the incentive layer changes the exploration–exploitation trade-off. The simulator and metric definitions are designed to support all four.

## 9. Conclusion

Long-lived personal agents have acquired the attributes of employment—identity, memory, tenure, obligations—without the institution that makes employment productive. We argued that the missing layer is an incentive mechanism: a contract over the agent's operating point that the agent's policy consumes. We formalized the mechanism, proved when it is real (non-consumption invariance), why persistence matters (the discount-factor threshold), how to trade capability against incentive intensity (the exchange-rate frontier), and how to bound gaming (the multi-channel Goodhart bound). We then specified PAIR, a runtime that implements the mechanism with an identity and tenure ledger, a compensation ledger, an authority ladder, a performance review, an audit scheduler, and a router, and we defined the metric suite that makes the mechanism observable.

The empirical contribution is deliberately scoped. In a controlled simulation of 3,000-task personal-agent workloads across 20 seeds, an incentive-responsive runtime improves ROI relative to static allocation and to scalar per-task rewards, while reducing the gaming rate. In a 240-run real-LLM experiment, prompt-level incentive framing has no detectable effect, whereas a runtime allocation policy raises verified success on the hardest task family from 0.60 to 1.00 at 4.8x price-weighted cost, with the exchange-rate theorem predicting the break-even baseline. The results are not claims about a specific product; they are evidence that the mechanism can be implemented, measured, and compared, and that its qualitative predictions survive contact with both a simulated and a real agent.

The larger point is about where the next unit of progress lies. Capability is becoming a purchasable input. What remains scarce is the discipline to allocate it: to decide which tasks deserve which agent, under what authority, with what verification, and with what consequences. Incentive engineering is that discipline. It is not a metaphor borrowed from management; it is the control layer that a persistent agent needs in order to keep acting well.

# Appendix A. Proofs

## A.1 Proposition 1

The agent solves max_a ∈ $\mathcal{A}\mathbb{E}$[u(a,ω)| I]. If $\mathcal{A}$(I)=$\mathcal{A}$(I') and u(·| I)=u(·| I'), the objective is pointwise identical, so the argmax set is identical. □

## A.2 Theorem 1

Consider the repeated game with per-period payoffs $u_H=wsp_H-c_H$ under high effort and $u_L=wsp_L-c_L$ under low effort, and grim-trigger reversion to the spot contract with bonus w=0 (per-period payoff zero) after a verified failure. Under cooperation the agent receives $u_H$ in every period, with value $u_H/(1-\delta)$. A one-shot deviation yields $u_L$ today and zero thereafter. High effort is sequentially rational iff

$$u_H/(1-\delta) \geq u_L \Leftrightarrow \delta \geq (u_L - u_H)/u_L. \quad \text{(A.1)}$$

Substituting the definitions and using $\Delta c=c_H-c_L$, $\Delta p=p_H-p_L$ gives (10). The max with zero covers the case in which the static incentive-compatibility condition $ws\Delta p\geq\Delta c$ already holds; then $u_H\geq u_L$ and $\delta^*=0$. The assumption $u_L>0$ ensures the denominator is positive and the spot contract is individually rational; if $u_L\leq 0$ the agent does not participate and the high-effort equilibrium is infeasible. □

## A.3 Theorem 2

The agent's incentive-compatibility constraint for high effort is $ws\Delta_m\geq c$, so the minimum bonus is $w_m=c/(s\Delta_m)$. Limited liability requires $w_m\leq v$, i.e. $v\geq c/(s\Delta_m)$. The principal's profit under high effort at tier m is

$$\Pi_H(m) = (v - w_m s)(p_m+\Delta_m) - k_m = (v - c/\Delta_m)(p_m+\Delta_m) - k_m, \quad \text{(A.2)}$$

and under low effort it is $v\, p_m-k_m$. High effort is profitable iff $\Pi_H(m)\geq v\, p_m-k_m$, which rearranges to $v\geq c(p_m+\Delta_m)/\Delta_m^2$. Taking the maximum with the limited-liability condition gives (11). Comparing $\Pi_H(S)$ with $\Pi_L(L)=v\, p_L-k_L$ gives

$$v(p_S+\Delta_S-p_L) \geq k_S - k_L + (c/\Delta_S)(p_S+\Delta_S), \quad \text{(A.3)}$$

and the maximum with the limited-liability condition gives (12). If $p_S+\Delta_S\leq p_L$ the denominator is non-positive and the large tier dominates at every v. □

## A.4 Theorem 3

Let a false claim be profitable only if it is not detected, and let the penalty when detected at least offset the gain. Then the expected profit of a false claim is positive only on the event that no channel detects it. Under independence, that event has probability $\Pi_k(1-s_k)$. In the long run, the fraction of claims that can be profitably false is at most the probability that a false claim is undetected, which is (14). Detection by at least one channel has probability $1-\Pi_k(1-s_k)$, which is (13). □

## A.5 Proposition 2

Write $J(x;\tau)=v\, p(x,e^*(x))-C(x;\tau)-r(a)$ on the finite grid X. Assume (i) p is nondecreasing and supermodular in (m,d), and nondecreasing in κ; (ii) p is nondecreasing in v (trivially, through the multiplicative term) and in the sense that $\partial^2 J/\partial x_i\partial v=\partial p/\partial x_i\geq 0$ for each capability lever $x_i$; (iii) C is increasing in m,d,ν and linear in λ with $\partial C/\partial\lambda=L(x)$, so $\partial^2 J/\partial x_i\partial\lambda=-\partial L/\partial x_i\leq 0$ for tiers and depth; (iv) authority a is nondecreasing in success probability when it gates a task, and the incident risk r(a) does not depend on v or λ. Then J has increasing differences in (x,v) and decreasing differences in (x,λ). By Topkis's theorem, the maximal and minimal selections of $\text{argmax}_{x\in}$ $_X$J(x;τ) are nondecreasing in ν and nonincreasing in λ. The claims for m,d,κ,ν follow by applying the single-crossing conditions componentwise; the claim for a follows from the gating assumption and the fact that r(a) is independent of (λ,ν). □

The conditions are sufficient, not necessary. The simulator's response model satisfies them by construction for the tier, depth, memory, and verification levers; authority has a non-monotone incident risk, which is why the router treats a as a constrained choice rather than relying on the proposition.

## Appendix B. Simulator parameters

The simulator is fully specified by `experiments/results/parameters.csv` (87 entries). Table B1 lists the parameters that determine the qualitative behaviour of the mechanism; the remaining entries are bookkeeping and protocol settings.

**Table B1.** Key PAW-Sim parameters.

| Parameter | Value | Role |
|---|---|---|
| Tasks per episode; warm-up; seeds | 3,000; 300; 20 | protocol |
| Complexity mix (main / stress) | 0.50/0.35/0.15; 0.20/0.40/0.40 | task stream |
| Task value base by complexity | 1 / 6 / 30, log-normal σ=0.4 | principal value |
| Latency price λ | U(0.04, 0.16) | cost |
| Verifiability q | U(0, 1) | task attribute |
| Tier capability θ | 0.0 / 0.8 / 1.6 (S/M/L) | success logit |
| Token price per 1k | 0.00015 / 0.0006 / 0.0024 | cost |
| Token count | (400 + 1200c)·d·(1+κ) | cost |
| Latency | $0.8 \cdot d \cdot \mu_m$, μ=(1,2,4) | cost |
| Verification cost ν=0/1/2 | 0 / 0.2 / 0.6 | cost |
| Effort response weights | depth 0.25, memory 0.50, effort 0.30, complexity −0.15, verification penalty −0.35 | success logit |
| Effort cost $c_e$ | 1.0, quadratic | agent |
| Gaming g; audit $a_{det}$ | σ(1.5 − 2.5νq); min(1, 0.5νq) | gaming |
| Bonus w | 0.5·v (R1 on claim; R2, R3 on verified success) | contract |
| Base salary fraction | 0 / 0 / 0.01 / 0.02 (R0 / R1 / R2 / R3) | contract |
| Discount δ | 0 / 0 / 0.2 / 0.5 | persistence |
| Decay time constant τ | 60 / 120 / 400 / 2,000 tasks | persistence |
| Intrinsic weight ψ | 0 / 0 / 0 / 0.3 | R3 only |
| Review period | 100 tasks | architecture |
| Promotion thresholds | 15 / 60 / 180 verified successes; reputation 0.50 / 0.55 / 0.60 | architecture |
| Authority levels; incident risk | 0–4; $0.002a^2$ | architecture |

The values marked “chosen” in `parameters.csv` are free modelling choices, not estimates from a deployed agent. They were fixed before the main run and are varied by ±50 percent in the sensitivity analysis (Figure 6).

## Appendix C. Additional results

Table C1 reports the stress condition (20/40/40). The ordering is unchanged: R3 has the highest ROI and the lowest gaming rate, and the unverified bonus in R1 again buys claims rather than work. The gap between R3 and R0 is somewhat smaller in relative terms because the stress mix contains more expensive tasks, which the router serves with the large tier.

**Table C1.** Stress condition (20/40/40 complexity mix; mean over 20 seeds, 2,700 evaluated tasks; 95% bootstrap CI in brackets).

| Regime | ROI | Verified success | Gaming rate | Net principal |
|---|---|---|---|---|
| R0 | 10.26 [10.07, 10.46] | 0.405 | 0.440 | 8,070 |
| R1 | 10.17 [10.00, 10.33] | 0.401 | 0.440 | −8,955 |
| R2 | 7.19 [7.05, 7.33] | 0.464 | 0.210 | −3,182 |
| R3 | 21.72 [21.57, 21.88] | 0.850 | 0.073 | 15,476 |

## Appendix D. Reproducibility and artifact inventory

The study is a deterministic simulation. The full pipeline is:

```
python3 experiments/sim/run_all.py --out experiments/results
python3 experiments/sim/make_figures.py --results experiments/results --figures experi-
ments/figures
python3 paper/build/build.py           # regenerates main.html and main.pdf
```

The first command writes `per_seed.csv`, `summary.json`, `decay.csv`, `frontier.csv`, `ablations.csv`, `sensitivity.csv`, `parameters.csv`, and `run.log`. The second writes the figures used in Section 7. The third rebuilds the PDF from the Markdown fragments in `paper/sections/` and the verified bibliography in `refs/refs.bib`.

Seeds are fixed and recorded per row of `per_seed.csv`. No randomness enters the analysis outside the simulator; confidence intervals are bootstrap resamples over the recorded seed×regime cells, so re-running the analysis on the released CSVs reproduces every table. The simulator uses only numpy and pandas; the paper build uses only Python's standard library and headless Chrome.

**Artifact list.** (i) simulator source; (ii) parameter file; (iii) all result CSVs and the summary JSON; (iv) figure sources and rendered figures; (v) this manuscript's Markdown sources, stylesheet, build script, and bibliography. The bibliography entries were resolved against the arXiv API and OpenAlex on 2026-11-06; each carries a `note = {verified 2026-11-06}` field.

**Tenure × verification interaction.** Theorem 1 predicts that identity continuity matters most when static incentive compatibility fails, i.e. when verification is weak. We ran the 2×2 interaction (tenure $\delta \in \{0, 0.5\}$ × verification $\nu$=0 or router) within R3. Mean effort is 0.492 ($\delta$=0, $\nu$=0), 0.503 ($\delta$=0.5, $\nu$=0), 0.529 ($\delta$=0, router), and 0.555 ($\delta$=0.5, router): tenure raises effort by 2.2 percent under weak verification and by 4.8 percent under routed verification, and net principal ROI moves from 0.583 to 0.574 (weak verification) and from 0.630 to 0.631 (routed). The interaction is therefore not the one the theorem predicts for a binding region: both cells remain in the region where the static condition $ws\Delta p \geq \Delta c$ holds, so $\delta^*$=0 and continuation value has little room to act. We report this as a boundary case, not as support for the theorem. Testing the theorem requires a calibration with weak verification *and* a bonus below the static threshold, which we did not run; we flag it as the most important follow-up experiment.

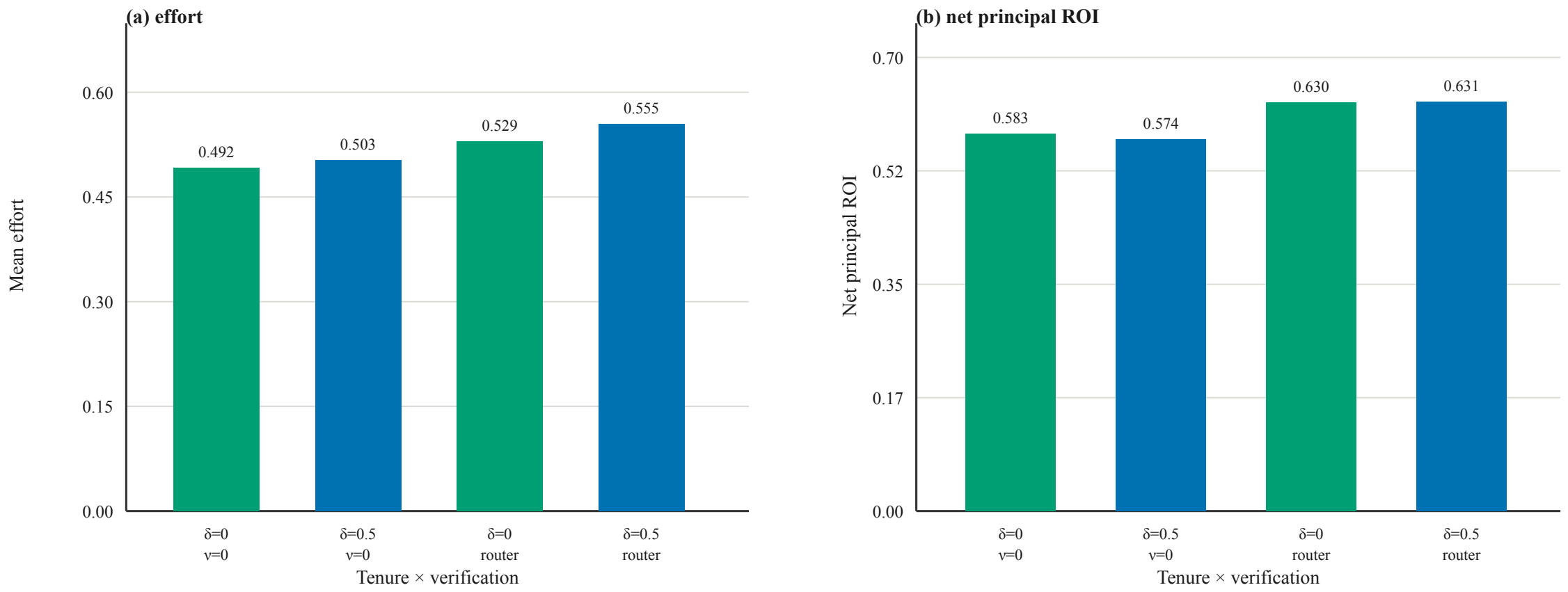

Figure C1. Tenure × verification interaction. Left: mean effort; right: net principal ROI. The tenure effect is small in both verification conditions, consistent with the calibrated region sitting above the persistence threshold (Theorem 1, $\delta^{*}=0$).